\documentclass[11pt]{article}

\PassOptionsToPackage{table}{xcolor}

\usepackage{xcolor}
\usepackage{acl}

\usepackage{booktabs}
\usepackage{times}
\usepackage{latexsym} 

\usepackage[T1]{fontenc}

\usepackage[utf8]{inputenc}

\usepackage{microtype}

\usepackage{inconsolata}

\usepackage{graphicx}

\usepackage{booktabs}
\usepackage{makecell}
\usepackage{multirow}
\usepackage{pifont}
\usepackage{tabularx}
\usepackage{amsmath}
\usepackage{amssymb}
\usepackage[ruled,vlined]{algorithm2e}
\usepackage{placeins}

\newcommand{\papername}{STP-GUI }
\title{STP-GUI: \underline{\textit{S}}patially-Aware \underline{\textit{T}}oken \underline{\textit{P}}runing for Efficient Web-\underline{\textit{GUI}} Agents}

\author{
\textbf{Shuo Wang\textsuperscript{1$\dagger$}},
 \textbf{Sifan Zhou\textsuperscript{1,2$\dagger$$\ddagger$}},
 \textbf{Xiaobo Wang\textsuperscript{3}}, 
 \textbf{Tianyu Fu\textsuperscript{1}}, \\
 \textbf{Chenxu Zhao\textsuperscript{1}}, 
 \textbf{Anyang Su\textsuperscript{1}}, 
 \textbf{Hanning Wang\textsuperscript{1}}, 
 \textbf{Ping Jiang*\textsuperscript{1}}, 
 \textbf{Minghui Wu*\textsuperscript{1}}
\\
 \textsuperscript{1}Mininglamp,
 \textsuperscript{2}Southeast University,
 \textsuperscript{3}Shenzhen University of Advanced Technology
}

\begin{document}
\maketitle

\def\thefootnote{$\dagger$}\footnotetext{Equal contribution}
\def\thefootnote{$\ddagger$}\footnotetext{Work done as an intern at Mininglamp}

\begin{abstract}
With the rapid development of vision-language models (VLMs), Web-GUI agents have achieved promising performance in interacting with real-world interfaces through screenshot-based grounding. However, high-resolution screenshots and multi-turn histories introduce excessive visual tokens, causing substantial inference overhead. Existing token pruning methods, mainly designed for natural images, often select tokens independently based on importance scores, which can fragment the local visual structures required for recognizing small GUI elements such as buttons, icons, and hyperlinks. In this work, we propose \textbf{\papername} (\textbf{S}patially-aware \textbf{T}oken \textbf{P}runing for GUI agents), a query-guided token pruning framework that preserves fine-grained instruction-conditioned token scoring while performing spatially coherent window-level selection under aggressive token reduction. A lightweight selector is optimized through policy learning while the VLM backbone remains frozen. Experiments on Online-Mind2Web show that \papername achieves a favorable efficiency-performance tradeoff under low token budgets. At 21\% token retention, it preserves 77.7\% of full-token performance while reducing inference latency by 31\% and FLOPs by 57\%. Results on Android in the Wild further suggest that the method transfers to mobile GUI environments.
\end{abstract}

\section{Introduction}
\label{sec:intro}

\begin{figure}[!t]
\centering
\includegraphics[width=0.85\linewidth]{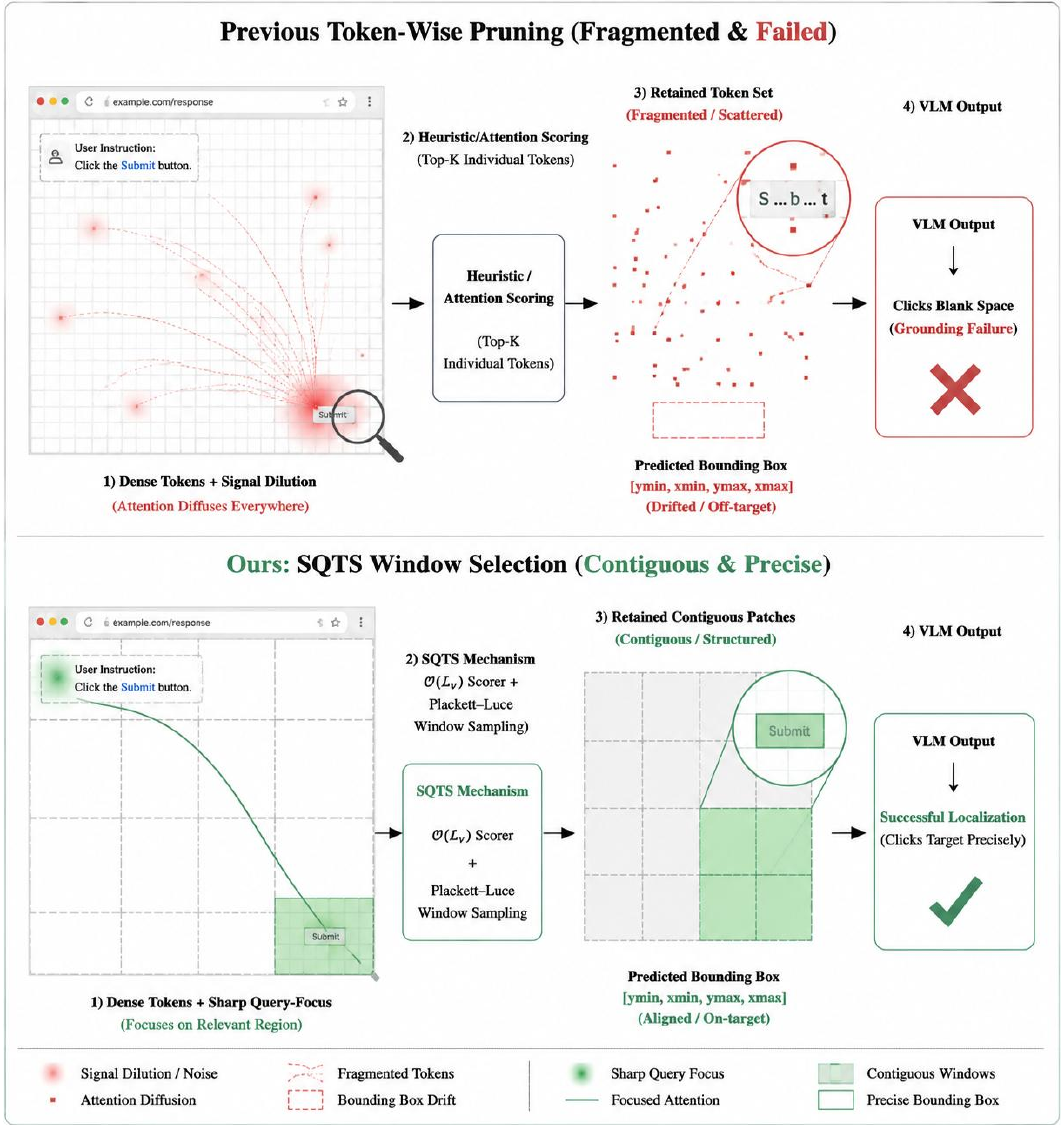}
\caption{Compared to existing VLM token pruning methods, our framework for Web‑GUI agents aggregates fine‑grained scores into non‑overlapping spatial windows and uses a Plackett–Luce policy during training to select complete, contiguous patches. This helps suppress background noise while preserving local semantic and geometric structure, enabling the frozen VLM backbone to retain action‑relevant evidence for downstream decisions.
}
\label{fig:teaser}
\vspace{-20pt}
\end{figure}

  
With the rapid development of vision-language models (VLMs), Web-GUI agents that navigate real websites by grounding actions in screenshots and text instructions have seen growing adoption~\cite{fu2025mano,wang2025ui,bai2025qwen3, bai2025qwen25vltechnicalreport, wang2024qwen2}. This setting requires high-resolution visual inputs to resolve small interactive elements such as hyperlinks, icons, and drop-down menus. The cost is a large visual token sequence: for instance, Qwen3-VL~\cite{bai2025qwen3} encodes a single 1920$\times$1080 screenshot into approximately 2,040 visual tokens. Furthermore, as is a common practice in Web-GUI agent systems~\cite{fu2025mano, qin2025uitarspioneeringautomatedgui,wang2025ui, bai2024digirl}, agents receive multiple interaction history---prior screenshots, actions, and observations---at each step, which can easily push the total visual token count beyond 10K. These large prefill token counts severely bottleneck inference, yet users are sensitive to agent response latency. Reducing visual token load is therefore a practical challenge.


A straightforward strategy is to prune visual tokens before input the LLM modules, as has been studied for VLM~\cite{shen2025fastvid,shang2025llava, yang2025visionzip}. Web-GUI screenshots, however, pose distinct challenges that make standard pruning strategies unreliable. First, informative regions are sparse, localized, and often tiny---a single icon or hyperlink may occupy only a handful of visual tokens---while most of the page consists of blank background, repeated layout blocks, or decorative structure. Second, the spatial continuity of these small regions is essential: losing a few adjacent patches can destroy the visual evidence needed to identify a clickable element. These characteristics make pruning strategies designed for natural images~\cite{liu2026hiprunehierarchicalattentionefficient,xing2024pyramiddrop}, where foreground objects are large and continuous, a poor fit for web interfaces. 

Despite these challenges, prior pruning paradigms in the VLM literature---attention-based selection~\cite{liu2026hiprunehierarchicalattentionefficient,xing2024pyramiddrop,shang2025llava}, similarity-based filtering~\cite{cai2025flashvlm,yang2025visionzip,jeddi2025similarity}, and learnable summarization~\cite{cha2024honeybee} may remain suboptimal for GUI screenshots. Existing pruning methods~\cite{liu2026hiprunehierarchicalattentionefficient,sun2026ivcprunerevealingimplicitvisual} involves two related
decisions: estimating instruction relevance and organizing
the retained evidence. Token-wise selection offers fine-grained relevance ranking,
but may retain disconnected fragments of interface elements.
Window-wise selection imposes local spatial structure,
yet its effectiveness also depends on whether the underlying
scores identify useful regions.
We therefore study scoring and selection granularity
jointly. Our matched-budget comparison shows that pooling Attention
scores into windows yields modest improvements, while
window-wise selection provides larger gains with the learned
STP-GUI scorer.
This motivates a lightweight instruction-conditioned
selector that estimates token relevance and retains complete
spatial windows before LLM processing. 


As illustrated in Fig.~\ref{fig:teaser}, we propose \textbf{STP-GUI} (\textbf{S}patially-aware \textbf{T}oken \textbf{P}runing for GUI agents), a query--guided token pruning method that selects spatially contiguous \emph{windows} rather than isolated tokens. Concretely, STP-GUI first computes token‑wise importance scores using a lightweight query--conditioned scoring module, aggregates the scores within $W{\times}W$ windows, and then uses a Plackett--Luce policy during training to learn which windows to retain. At inference time, windows are selected greedily by score rank under a fixed token budget, while the VLM backbone remains frozen throughout. This design preserves fine‑grained token scoring while reducing spatial fragmentation of the retained set, thereby helping the model preserve local evidence around GUI elements such as buttons, icons, and label‑field pairs. Experiments on Online-Mind2Web (300 tasks across 136 live websites)~\cite{xue2025illusion} show that STP-GUI is particularly competitive at low retention budgets. By evaluating on Android in the Wild (AitW)~\cite{rawles2023androidwildlargescaledataset}, we further show that the same design transfers beyond web screenshots to mobile GUI settings. At 21\% token retention, it achieves the strongest success rate among the compared methods reported at their evaluated operating points while reducing inference latency by 31\% and FLOPs by 57\% relative to the dense baseline. Our main contributions are summarized as:
\begin{itemize}

    \item We identify a previously overlooked failure mode of GUI-agent token pruning: coupling importance estimation and selection granularity causes spatial fragmentation of action-relevant visual evidence. 
    
    \item We introduce STP-GUI, a granularity-decoupled selection paradigm that scores tokens precisely while selecting contiguous windows to preserve spatial coherence.


    \item We formulate GUI visual token retention as a budget-constrained discrete selection problem and design a lightweight query-conditioned selector trained with policy learning, while keeping the VLM backbone frozen.

    \item Experiments on Online-Mind2Web show that STP-GUI provides a favorable tradeoff between accuracy and efficiency under token pruning, reducing inference latency by 31\% and FLOPs by 57\%. Results on AitW also show the generalization for mobile GUI environments.

\end{itemize}

\vspace{-4mm}
\section{Related Work}

\paragraph{Web-GUI Agents.}
Early GUI agents relied on modular pipelines combining text extraction, reasoning, and memory components~\cite{hurst2024gpt-4o,xie2024osworld}, which can be brittle under changing environments~\cite{xia2024agentless}. Recent VLM-based agents instead directly ground actions from screenshots and achieve stronger generalization on benchmarks such as Mind2Web and AitW~\cite{wang2024cogvlm,hong2024cogagent,chen2025guicourse}. Meanwhile, mobile GUI agents further improve grounding through specialized training pipelines~\cite{tang2025magicgui}. However, regardless of agent architecture, high-resolution visual inputs introduce substantial token overhead, making efficient visual token selection an important problem for deployment.

\paragraph{Token Pruning in VLMs.}
Token pruning reduces the inference cost of VLMs by removing redundant visual tokens. Existing approaches can be broadly categorized into training-based and training-free methods~\cite{bolya2022token}. 
Training-based methods, such as ATP-LLaVA~\cite{ye2025atp}, learn adaptive pruning policies or ratios, while STP-GUI instead freezes the VLM backbone and optimizes a lightweight query-conditioned selector for spatially coherent pruning.
Training-free methods mainly rely on importance-based scoring (e.g., FastV~\cite{chen2024image}, SparseVLM~\cite{zhang2024sparsevlm}, MustDrop~\cite{liu2024multi}, PDrop~\cite{xing2024pyramiddrop}) or redundancy-based filtering (e.g., DART~\cite{wen2025stop}, DivPrune~\cite{alvar2025divprune}). 
However, these methods are primarily designed for natural images and select tokens independently, which may fragment the spatial evidence required for GUI grounding. STP-GUI addresses this limitation by decoupling token relevance estimation from evidence retention granularity, enabling spatially coherent token selection under constrained budgets.

\begin{figure*}[t]
    \centering
    \includegraphics[width=0.8\textwidth]{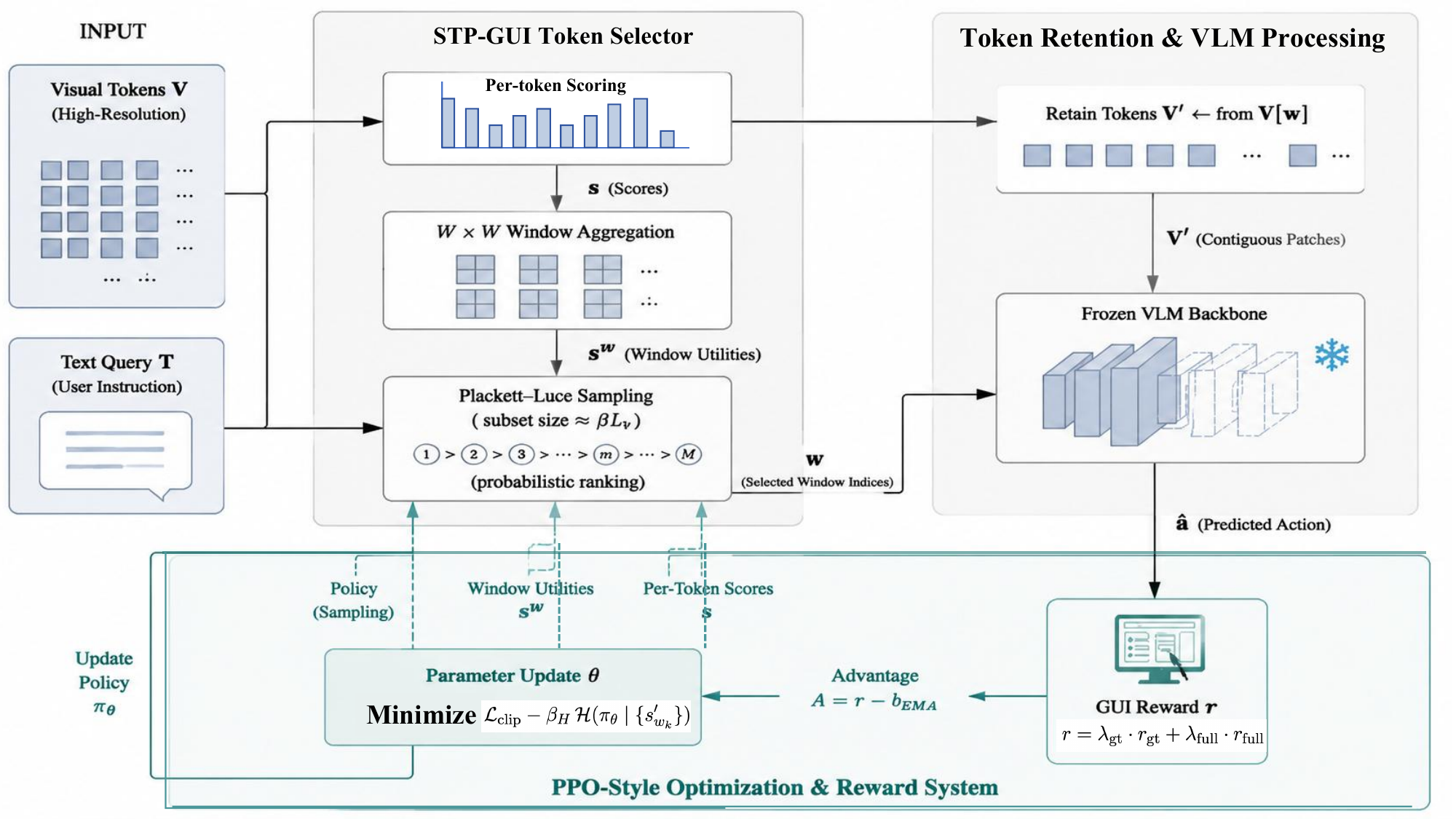} 
    \caption{\textbf{Overview of the \papername pipeline.} Given the task instruction and visual tokens, \papername scores token relevance and performs window-level selection under a token budget before the retained tokens are fed into the LLM. The resulting compact visual representation is then used for downstream action prediction.}
    \label{fig:pipeline}
    \vspace{-5mm}
\end{figure*}

\paragraph{Visual Grounding for GUI Agents.}
Recent works improve GUI grounding through richer supervision or region-aware representations. For example, Gao et al.~\cite{gao2024enhancing} enhance vision-language alignment with region-level supervision, YORO~\cite{ho2022yoro} develops efficient visual grounding models, and R-VLM~\cite{park2025rvlm} introduces region-aware visual representations for GUI grounding. 
Different from these approaches, STP-GUI addresses visual token reduction for a frozen GUI agent rather than modifying the grounding backbone. Its selector uses action-level supervision and/or
full-token model feedback, without requiring
token-retention masks or UI-element bounding boxes.


\vspace{-2mm}
\section{Methodology}
\vspace{-2mm}
\subsection{Preliminary}

A VLM processes visual tokens $\mathbf{V} \in \mathbb{R}^{L_v \times d}$ concatenated with text tokens $\mathbf{T} \in \mathbb{R}^{L_t \times d}$. Since $L_v \gg L_t$, token pruning selects a subset of $\mathbf{V}$ via a scoring policy $\mathcal{P}$:
\begin{equation}
\mathbf{S} = \mathcal{P}(\mathbf{V}, \mathbf{T}), \quad
\mathbf{M}_i = \mathbf{1}\!\left[\mathbf{S}_i \in \mathrm{Top}\text{-}N(\mathbf{S})\right],
\end{equation}
where $L_v$ and $L_t$ denote the number of visual and text tokens respectively, and $d$ is the feature dimension. $\mathbf{M} \in \{0,1\}^{L_v}$ is the binary retention mask and $N = \lfloor \beta L_v \rfloor$ is a fixed retention budget. We refer to $\beta$ as the \emph{keep ratio}, i.e., the fraction of visual tokens retained after pruning. For multi-turn GUI agent interaction where input history concatenated with current observation yields multiple image frames, STP-GUI applies the window selection mechanism frame-wise under a uniform keep ratio $\beta$. This ensures that each temporal visual step retains a spatially contiguous and coherent local representation before entering the LLM.


\vspace{-2mm}
\subsection{Method Overview}
\label{sec:bottleneck}



STP-GUI prunes adapter-projected visual tokens before they enter the LLM. Given visual tokens and an instruction, a lightweight
scorer produces token-level relevance scores.
These scores are aggregated within non-overlapping
spatial windows.
During training, windows are sampled under a fixed
budget and the selector is optimized using action-level
feedback from the frozen VLM.
At inference, the highest-scoring windows are retained
deterministically.



\vspace{-2mm}
\subsection{Instruction-Guided Token Scoring}
\label{sec:scorer}

We define a lightweight instruction-conditioned scorer $f_\theta(\mathbf{V}, \mathbf{T}): \mathbb{R}^{L_v \times d} \times \mathbb{R}^{L_t \times d} \rightarrow \mathbb{R}^{L_v}$ that maps visual tokens $\mathbf{V}$ and text tokens $\mathbf{T}$ to per-token importance scores $s_i = f_\theta(\mathbf{V}, \mathbf{T})_i$. $\mathbf{V}$ denotes the adapter-processed visual token sequence from the vision encoder, and $\mathbf{T}$ denotes the embedding sequence of the current instruction. Following~\cite{feng2025towards}, $\mathbf{T}$ is restricted to the action-relevant portion of the instruction, excluding chain-of-thought reasoning steps.

The scorer performs lightweight cross-modal interaction between projected visual and textual features, conditioning each visual token on the instruction to estimate its task relevance. The resulting per-token scores are subsequently aggregated for window-level selection.

STP-GUI shifts the selection unit from individual tokens to spatial windows while keeping token-level scoring for fine-grained importance estimation. The two operations are decoupled: token scores provide local precision, and window-level selection provides spatial coherence.

Specifically, given visual tokens on a spatial grid of size $H_g \times W_g$ (where $H_g$ and $W_g$ are the grid height and width in tokens, respectively, with $H_g \cdot W_g = L_v$), we partition the grid into non-overlapping spatial windows of size $W \times W$, indexed by $w \in \{1, \ldots, \lfloor H_g/W \rfloor \cdot \lfloor W_g/W \rfloor\}$. The per-token scores within each window are aggregated into a window-level score:
\vspace{-2mm}
\begin{equation}
\begin{aligned}
s_w = \phi(\{s_i\}_{i \in w})
\end{aligned}
\end{equation}
where $\phi(\cdot)$ aggregates the token importance scores within window $w$. These aggregated window scores parameterize a Plackett--Luce distribution over candidate windows. During training, we sample window subsets from this distribution so that the expected token budget matches the target keep ratio $\beta$, and optimize the resulting policy with PPO. At inference time, no policy optimization is involved: windows are ranked by the aggregated scores produced by the trained scorer $f_\theta$, and the top-ranked windows are selected under the same budget. Because every retained region is a complete $W \times W$ patch, the downstream VLM receives local evidence instead of isolated token fragments.

\vspace{-2mm}
\subsection{Training the Window Selector}
\label{sec:ppo}


We do not have ground-truth token-retention masks.
We therefore optimize the discrete window-selection
policy using action-level feedback from the frozen VLM. We therefore sample window subsets with a Plackett--Luce policy induced by the aggregated window scores, and use PPO to train the selector toward window choices that produce action responses consistent with the full-token model while respecting the token budget.

\paragraph{Reward design.}
We use two complementary reward signals: $r_{\mathrm{gt}}$, which encourages actions consistent with task-level supervision when available, and $r_{\mathrm{full}}$, which encourages the pruned-input policy to remain consistent with the action preference induced by the same model under the full, unpruned input. In other words, $r_{\mathrm{full}}$ pushes the selector to retain window subsets that elicit action responses close to those of the full-token model. 

\vspace{-2mm}
\begin{equation}
\begin{aligned}
r_{\mathrm{step}} = \mathrm{T_{mt}} \bigl( \lambda_{\mathrm{type}} + (1 - \lambda_{\mathrm{type}}) \cdot \mathrm{sim} \bigr)
\end{aligned}
\end{equation}

\noindent where $\mathrm{T_{mt}} \in \{0,1\}$ indicates the binary type match and $\mathrm{sim} \in [0,1]$ is the similarity score measured by edit distance for text arguments and tolerance-based matching for spatial arguments. For actions without parameters, we set $\lambda_{\mathrm{type}}=1$.

The final reward is a weighted combination of the two, allowing the selector to balance task supervision and full-context consistency:
\vspace{-2mm}
\begin{equation}
\begin{aligned}
r = \lambda_{\mathrm{gt}} \cdot r_{\mathrm{gt}} + \lambda_{\mathrm{full}} \cdot r_{\mathrm{full}}
\label{eq:reward}
\end{aligned}
\end{equation}
  where $\lambda_{\mathrm{gt}}, \lambda_{\mathrm{full}} \in [0, 1] $. When task-level supervision is unavailable, training can still rely on $r_{\mathrm{full}}$ alone. As shown in the appendix, either reward alone yields competitive results, the combined reward achieves the best overall performance.

\paragraph{Optimization.}
We maintain per-action-type EMA baselines for variance reduction and train the selector using a PPO-style clipped objective with entropy regularization:

\vspace{-2mm}

\begin{align}
\mathcal{L}_{\mathrm{clip}} &= -\mathbb{E}\!\left[\min\!\left(\rho A,\;\mathrm{clip}(\rho,\,1{-}\epsilon,\,1{+}\epsilon)\,A\right)\right], \label{eq:clip_loss} \\
\mathcal{L} &= \mathcal{L}_{\mathrm{clip}} - \beta_H\,\mathcal{H}(\pi_\theta \mid \{s_{w_k}\}), \label{eq:total_loss}
\end{align}

where $\rho = \pi_\theta / \pi_{\mathrm{old}}$ is the policy ratio, $A = r - b_{\mathrm{ema}}$ is the advantage with $b_{\mathrm{ema}}$ being the per-action-type EMA baseline, $\epsilon$ is the clipping threshold, and $\beta_H$ controls the entropy regularization strength. \(\mathcal{H}(\pi_\theta \mid \{s_{w_k}\})\) denotes the entropy of the resulting window-selection policy. The entropy term is used to encourage exploration during training. The VLM backbone is frozen and only the lightweight selector is updated. We optimize the discrete selection policy using PPO
with entropy regularization.
PPO is an optimization choice rather than a separate
novel algorithmic component.
Nevertheless, the optimizer affects the achieved
performance, as evaluated in
Table~\ref{tab:ppo_vs_reinforce}.


\vspace{-1mm}

\vspace{-2mm}
\section{Experiments}

\paragraph{Benchmarks.}
\textbf{Online-Mind2Web}~\cite{xue2025illusion} is a live web agent benchmark comprising 300 tasks across 136 real websites, using task-level success rate (SR) as the primary metric. Tasks require end-to-end interaction with live pages, making them sensitive to spatial-localization errors introduced by aggressive token pruning.
  \textbf{Android in the Wild (AitW)}~\cite{rawles2023androidwildlargescaledataset} is a large-scale mobile GUI dataset spanning diverse app categories. Following \cite{bai2024digirl}, we evaluate on the General and Web Shopping test splits to assess cross-domain transferability of STP-GUI.

\vspace{-2mm}
\paragraph{Baseline model construction.}
We used Qwen3-VL as the backbone model for our experiments. We construct the Web-GUI supervised training data following prior work~\cite{qin2025uitarspioneeringautomatedgui, wang2025ui, fu2025mano}. For Online-Mind2Web and AitW, we separately obtain the corresponding backbones by supervised fine-tuning(SFT) Qwen3-VL on their training data. STP-GUI and all pruning baselines are evaluated on top of these SFT backbones. We only performed SFT as the baseline and developing advanced RL method to achieve a higher accuracy is beyond the scope of this paper.

\paragraph{Implementation details.}
We fine-tune Qwen3-VL as the base model using LLaMA-Factory~\cite{zheng2024llamafactory} on 8$\times$A100 (80G) GPUs for 3 epochs with an initial learning rate of $1\text{e-}5$ and a cosine decay schedule. During SFT, the ViT encoder is frozen, while the adapter and LLM parameters are updated. After SFT, the VLM backbone is completely frozen for token pruning. STP-GUI only optimizes a lightweight 2M-parameter selector for window retention, without updating any VLM parameters. Unlike pruning methods that adapt VLM parameters during optimization, STP-GUI introduces only a small selection module that can be trained offline and deployed on top of the frozen backbone. The detailed tuning cost comparison with adapted pruning baselines is provided in appendix.
Training the selector takes approximately 10 hours on 8 GPUs. This additional cost is incurred only once offline, while the learned selector reduces inference latency by 31\% and FLOPs by 57\% during deployment. 

with entropy regularization.
PPO is an optimization choice rather than a separate
novel algorithmic component.
Nevertheless, the optimizer affects the achieved
performance, as evaluated in Table~\ref{tab:ppo_vs_reinforce}.

For comparison methods, we re-implement representative pruning approaches~\cite{xing2024pyramiddrop,liu2026hiprunehierarchicalattentionefficient,sun2026ivcprunerevealingimplicitvisual,cai2025flashvlm} on Qwen3-VL using their official implementations. To ensure fair comparison, we evaluate both their original training-free settings and fine-tuned variants adapted on the same SFT data. Layer-sensitive baselines are evaluated using their best-performing configurations, with detailed sensitivity analyses provided in the appendix. Since Online-Mind2Web is evaluated on live websites, we repeat experiments 3 times under the same protocol and report the mean SR. 





\begin{table}[t]
    \centering
    \small
    \caption{Task success rate (SR) on Online-Mind2Web. \textit{Keep ratio} (KR) denotes the fraction of visual tokens retained. The baseline means without token pruning. Best results are in \textbf{bold}. * denotes our re-training implementation for training-free methods. ACC represents the ratio relative to the full-token model. Partial baselines are reported at their native setting, while direct same-ratio comparisons are available where matched keep ratios are reported.}
    \label{tab:performance_comparison}
    \begin{tabular}{llccc}
        \toprule
        Model & Method & KR  & SR ($\uparrow$) & ACC($\uparrow$) \\
        \midrule
        \multirow{17}{*}[0pt]{\rotatebox[origin=c]{90}{Qwen3-VL-2B}}
            & Baseline  & 100\% & 0.373 &100\%\\
            \cmidrule{2-5}
      & \multirow{2}{*}{HiPrune}  & 21\% & 0.100  & 26.8\% \\
            &                           & 50\% & 0.183 & 49.1\% \\
            & \multirow{2}{*}{HiPrune*}  & 21\% & 0.140 & 37.5\% \\
            &                           & 50\% & 0.193 & 51.7\% \\ \cmidrule{2-5}
            & PDrop    & 41\% & 0.120 & 32.2\% \\
       & PDrop*    & 41\% & 0.143 & 38.3\% \\
      \cmidrule{2-5}
            & IVC & 50\% & 0.206 & 55.2\%\\
            & IVC* & 50\% & 0.217 & 58.2\%\\ \cmidrule{2-5}
      & \multirow{2}{*}{FlashVLM} & 21\% & 0.153 & 41.0\%\\
            &                           & 50\% & 0.177 & 47.5\%\\
            & \multirow{2}{*}{FlashVLM*} & 21\% & 0.157 & 42.1\% \\
            &                           & 50\% & 0.263 & 70.5\% \\ \cmidrule{2-5}
            & \multirow{2}{*}{Attention} & 21\% & 0.220 & 58.9\% \\
            &                            & 50\% & 0.280 & 75.1\% \\
            \cmidrule{2-5}
            & \multirow{2}{*}{\textbf{STP-GUI}} & 21\% & \textbf{0.290} & 77.7\% \\
            &                                & 50 \% & \textbf{0.350} & \textbf{93.8}\% \\
        \midrule
        \multirow{8}{*}[0pt]{\rotatebox[origin=c]{90}{Qwen3-VL-4B}}
            & Baseline & 100\% & 0.403 & 100\% \\
            \cmidrule{2-5}
      & IVC* & 50\% & 0.246 & 61.0\% \\
      \cmidrule{2-5}
      & FlashVLM & 50\% & 0.213 & 52.9\%  \\
      & FlashVLM* & 50\% & 0.27 & 72.4\% \\
      \cmidrule{2-5}
      
      &\multirow{2}{*}
      {Attention} & 21\% & 0.227 & 56.3\% \\
      & &50\%  & 0.303 & 75.0\% \\ \cmidrule{2-5}
            &\multirow{2}{*}{\textbf{STP-GUI}} & 21\% & \textbf{0.307} & 76.2\%\\
      & &50\%  & \textbf{0.383} & \textbf{95.0}\% \\
        \bottomrule
    \end{tabular}
    \vspace{-7mm}
\end{table}

\paragraph{Comparison with Other Pruning Methods.}

Tab.~\ref{tab:performance_comparison} compares STP-GUI against representative visual token pruning methods on Online-Mind2Web. We evaluate 2 keep ratios (21\% and 50\%) on Qwen3-VL-2B, and 21\% and 50\% on Qwen3-VL-4B where reported. ``Attention'' selects tokens by prompt-to-image cross-attention scores averaged over LLM decoder layers. Following the layer sensitivity analysis in the appendix, we adopt layer 16 for Attention, which yields the strongest results within that method family. At 21\% retention, STP-GUI reaches SR\,=\,0.290, where it can be directly compared against HiPrune, HiPrune*, FlashVLM, FlashVLM*, and Attention at the same keep ratio. Notably, this result already exceeds PDrop and PDrop*, which are reported at their native operating point of about 41\%, as well as IVC and IVC*, which are reported at 50\% in our reproduction. At 50\%, STP-GUI further reaches 0.350, remaining ahead of the other available methods at the same operating point. Overall, the table shows that STP-GUI remains competitive under substantially tighter token budgets, while FlashVLM and HiPrune further incur substantial inference-time memory overhead. STP-GUI stays much closer to the dense baseline while being more memory efficient and lower latency.

\begin{table}[h]
\vspace{-2mm}
    \centering
    \small
    \caption{Results on AitW. Following \cite{bai2024digirl, feng2025towards}, we report General and Web Shopping splits.}
    \vspace{-2mm}
    \label{tab:aitw}
    \begin{tabular}{llcc}
        \toprule
        Method & KR & General & Web Shopping\\
        \midrule
            Baseline   & 100\%     & 0.782 & 0.525 \\ 
            \midrule
      FlashVLM*   & 50\%     & 0.176 & 0.081 \\
            PDrop*   & 41\%     & 0.178 & 0.131 \\
            HiPrune* & 50\% & 0.252 & 0.214\\
      Attention  & 50\%       & 0.338 & 0.338 \\
      \cmidrule{1-4}
            \multirow{2}{*}{\textbf{STP-GUI}} & 21\% & 0.718 & 0.436 \\
            & 50\% & 0.741 & 0.492\\
        \bottomrule
    \end{tabular}
    \vspace{-6mm}
\end{table}

\vspace{-2mm}
\paragraph{Generalization to AitW Dataset.}
To examine whether the benefit of STP-GUI extends beyond Online-Mind2Web, we additionally evaluate it on AitW, a mobile UI grounding benchmark with different interface layouts, visual styles, and action distributions. Following prior work~\cite{bai2024digirl, feng2025towards}, we report results on the General and Web Shopping splits. As shown in Tab.~\ref{tab:aitw}, all compared methods are evaluated on top of backbones separately fine-tuned on the AitW training data, and we keep the pruning configurations consistent with those used in the main experiments. For PDrop, the retained token ratio is determined by its progressive halving design across 3 layers. Under the same 50\% budget, STP-GUI outperforms Attention by a large margin on both splits (0.741 vs.\ 0.338 on General, 0.492 vs.\ 0.338 on Web Shopping), showing that the advantage of STP-GUI is not limited to the web setting. At the more aggressive 21\% keep ratio, STP-GUI still maintains strong transfer performance, achieving 0.718 on General and 0.436 on Web Shopping. Because not all baselines are evaluated at exactly the same keep ratio, we do not claim a strict iso-budget comparison for every entry in Tab.~\ref{tab:aitw}. Rather, these results show that STP-GUI remains competitive when the token budget is tightened substantially below the operating points used by several competing methods.

Note that the different keep ratios reported in Tab.\ref{tab:performance_comparison} and Tab.\ref{tab:aitw} follow the original evaluation setting of the compared pruning methods, rather than being intentionally chosen to favor STP-GUI. Different pruning approaches adopt different native pruning schedules. For example, PDrop uses a progressive halving strategy, leading to a default retention ratio of approximately 41\%, whereas FlashVLM, HiPrune, IVC and STP-GUI are evaluated under explicitly specified retention budgets. Following the original implementations allows us to faithfully reproduce each method.

To further check whether the observed trend is not limited to the base backbone used in our main experiments, we additionally evaluate STP-GUI on a larger backbone, Qwen3.5-9B~\cite{qwen3.5}. We keep these results in the appendix, where they show a similar efficiency--performance trend on both Online-Mind2Web and AitW at a 50\% token budget.

\FloatBarrier



\vspace{-2mm}
\paragraph{Analysis of Spatial Coherence.} Beyond task success rate, we further analyze the spatial structure of the retained tokens. This analysis is motivated by the window-size ablation in \ref{fig:window_size_ablation}: when $W{=}1$ degenerates to token-wise selection, performance drops at the same keep ratio, suggesting that how retained tokens are organized in space may matter. Ideally, one would also measure whether the retained tokens continue to cover the action-critical target region at each interaction step. However, such a dataset-level evaluation is not directly available in either Online-Mind2Web or AitW: these benchmarks do not provide dense per-step UI-element annotations for complete trajectories, and successful interaction paths are not unique. We therefore focus on a trajectory-level analysis of the binary retention masks produced by different pruning methods at the same 21\% budget, using 4 connected-component statistics (8-connected-component statistics are shown in appendix) as a quantitative measure of spatial fragmentation, and complement this with manually curated case studies in the appendix, where we annotate step-level targets for representative successful Online-Mind2Web episodes and compare them with the retained token regions.

$N_{cc}$ is the number of connected components. A smaller value means less fragmentation.
$R_{\mathrm{largest}}$ is the fraction of retained tokens that belong to the largest connected component. A larger value means that most retained evidence stays in one dominant region.
$A_{\mathrm{mean}}$ is the mean area of connected components (in tokens). A larger value means the retained regions are larger and more complete.
$H_{cc}$ is the entropy of component sizes. A smaller value means the retained tokens are less evenly scattered across many small regions.
In short, better spatial coherence corresponds to lower $N_{cc}$ and $H_{cc}$, and higher $R_{\mathrm{largest}}$ and $A_{\mathrm{mean}}$.

\begin{table}[!htbp]
    \vspace{-2mm}
    \centering
    \scriptsize 
    \caption{Spatial coherence of retained token on Mind2Web at 21\% budget. $N_{cc}$: \#connected components; $R_{\mathrm{largest}}$: ratio of tokens in the largest component; $A_{\mathrm{mean}}$: mean component area; $H_{cc}$: component-size entropy. Lower $N_{cc}$/$H_{cc}$ and higher $R_{\mathrm{largest}}$/$A_{\mathrm{mean}}$ indicate more coherent retention.}
    \label{tab:spatial_coherence}
    \setlength{\tabcolsep}{2.5pt}
    \begin{tabular}{l c c c c}
        \toprule
        Method & $N_{cc}\downarrow$ & $R_{\mathrm{largest}}\uparrow$ & $A_{\mathrm{mean}}\uparrow$ & $H_{cc}\downarrow$ \\
        \midrule
        FlashVLM  & 124.56 $\pm$ 16.55 & 0.176 $\pm$ 0.059 & 3.24 $\pm$ 0.49  & 3.885 $\pm$ 0.343 \\
        HiPrune   & 128.95 $\pm$ 11.80 & 0.083 $\pm$ 0.030 & 3.10 $\pm$ 0.27  & 4.293 $\pm$ 0.179 \\
        Attention &  94.95 $\pm$ 17.68 & 0.272 $\pm$ 0.118 & 4.47 $\pm$ 0.89  & 3.234 $\pm$ 0.475 \\
        \midrule
        \textbf{STP-GUI} & \textbf{5.71 $\pm$ 2.29} & \textbf{0.780 $\pm$ 0.180} & \textbf{125.1 $\pm$ 76.1} & \textbf{0.724 $\pm$ 0.461} \\
        \bottomrule
    \end{tabular}
    \vspace{-2mm}
\end{table}


Tab.~\ref{tab:spatial_coherence} shows that STP-GUI produces only 5.71 connected components on average, compared to 94.95 for Attention and over 120 for FlashVLM and HiPrune. More than three quarters of its retained tokens (78\%) fall within a single dominant region, and the mean component area is 28--40$\times$ larger. In other words, competing methods scatter retained tokens into many small islands, while STP-GUI keeps them in one or a few coherent patches. These statistics describe the spatial organization
of retained masks.
Lower fragmentation is partly expected from
whole-window selection and does not independently
establish improved action-target coverage.
We interpret these results alongside the
task-performance comparisons in
Table~\ref{tab:scorer_window}.
The qualitative examples illustrate retained local
evidence rather than providing a dataset-wide measure
of target coverage.


\FloatBarrier
\vspace{-2mm}
\subsection{Ablation Studies}
\vspace{-2mm}
\begin{table}[!htbp]
    \centering
    \small
    \setlength{\tabcolsep}{4pt}
    \caption{
        Effect of selection granularity on Online-Mind2Web
        with Qwen3-VL-2B.
        KR denotes the visual token keep ratio.
        We report task success rate (SR) for token-wise
        selection ($W=1$) and window-wise selection ($W=4$).
        $\Delta$ denotes the SR improvement in percentage
        points. Attention uses scores from layer 16.
        The best SR at each keep ratio is bolded.
    }
    \label{tab:scorer_window}
    \begin{tabular}{@{}llccc@{}}
        \toprule
        KR & Method & $W=1$ & $W=4$ & $\Delta$ (pp) \\
        \midrule
        21\% & Attention & 0.220 & 0.230 & +1.0 \\
             & STP-GUI   & 0.247 & \textbf{0.290} & +4.3 \\
        \midrule
        50\% & Attention & 0.280 & 0.297 & +1.7 \\
             & STP-GUI   & 0.313 & \textbf{0.350} & +3.7 \\
        \bottomrule
    \end{tabular}
    \vspace{-5mm}
\end{table}

\paragraph{Scoring and selection granularity.}
To examine whether simple window aggregation is sufficient
to explain the performance of STP-GUI, we compare token-wise selection ($W=1$) and window-wise
selection ($W=4$) for Attention and STP-GUI at matched
keep ratios.
For Attention, scores from layer 16 are aggregated within
non-overlapping $4 \times 4$ windows, followed by greedy
selection under the token budget.
Table~\ref{tab:scorer_window} shows that window aggregation
improves Attention SR by 1.0 and 1.7 percentage points
at 21\% and 50\% retention, respectively.
STP-GUI achieves larger improvements of 4.3 and 3.7
percentage points when switching from token-wise to
window-wise selection.
With the same window size, STP-GUI exceeds Attention
by 6.0 and 5.3 percentage points.
These results suggest that window aggregation alone
is insufficient to explain the performance of STP-GUI,
and support the complementary roles of learned scoring
and spatially structured retention.



\begin{figure}[h]
    \vspace{-3mm}
    \centering
    \includegraphics[width=0.8\linewidth]{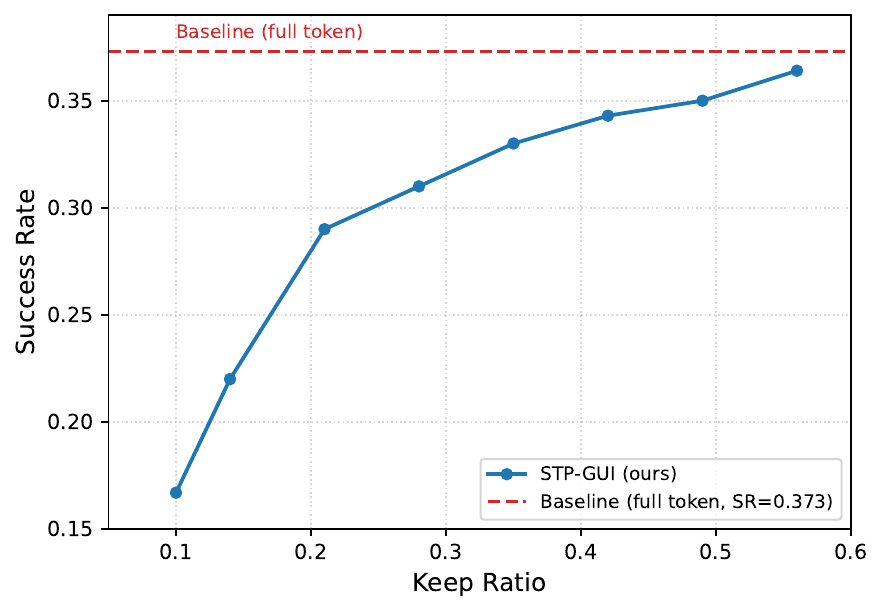}
    \vspace{-4mm}
    \caption{SR vs.\ KR on Online-Mind2Web. The dashed red line denotes the full-token baseline. STP-GUI achieves 77.7\% of baseline performance at KR = 0.21.}
    \label{fig:keep_ratio_ablation}
    \vspace{-4.2mm}
\end{figure}

\vspace{-2mm}
\paragraph{Effect of keep ratio.}
Fig.~\ref{fig:keep_ratio_ablation} shows SR across KR from 0.10 to 0.56, with the full-token baseline as reference. SR rises monotonically with the budget but exhibits diminishing returns: gains are steep at low ratios and flatten as the ratio approaches the baseline. At KR = 0.21, STP-GUI achieves SR\,=\,0.290, recovering 77.7\% full-token setting.


\begin{table}[!htbp]
\vspace{-2.5mm}
\centering
\small
\setlength{\tabcolsep}{8pt}
\caption{Effect of window overlap on task success rate.}
\vspace{-2mm}
\label{tab:overlap}
\begin{tabular}{lccc}
\toprule
Window strategy & Mind2Web & \makecell{AitW\\General}. & \makecell{AitW \\WebShop} \\
\midrule
Overlap ($W/2$) & 0.283 & 0.711 & 0.418 \\
Non-overlap & \textbf{0.290} & \textbf{0.718} & \textbf{0.436} \\
\bottomrule
\end{tabular}
\vspace{-5.5mm}
\end{table}

\paragraph{Effect of Window Overlap.}
We compare non-overlapping windows with an overlapping
variant using stride $W/2$ at $W=4$ and 21\% retention.
The overlapping variant does not improve the reported
scores on the three evaluation splits.
We therefore use non-overlapping windows as the default.
The small differences do not establish that boundary
coverage is unimportant.



\begin{table}[!htbp]
\vspace{-2mm}
\centering
\small
\caption{Comparison between PPO and naive REINFORCE with EMA baseline on Online-Mind2Web.}
\vspace{-3mm}
\label{tab:ppo_vs_reinforce}
\setlength{\tabcolsep}{12pt}
\begin{tabular}{lc}
    \toprule
    Optimizer & SR ($\uparrow$) \\
    \midrule
    Naive REINFORCE & 0.247 \\
    PPO (clipped objective)  & \textbf{0.290}  \\
    \bottomrule
\end{tabular}
\vspace{-4mm}
\end{table}

\paragraph{PPO vs.\ Simple REINFORCE.} The method section adopts a PPO-style clipped objective for training the window selector. A natural question is whether the clipping mechanism is strictly necessary or whether a simpler policy-gradient method (naive REINFORCE) would suffice. We compare the 2 optimization strategies on Online-Mind2Web with Qwen3-VL-2B at KR = 0.21. As shown in Tab.~\ref{tab:ppo_vs_reinforce}, PPO improves SR from 0.247 to 0.290, a gain of
4.3 percentage points.
This result shows that the optimization strategy
affects the achieved performance.
Additional training-seed statistics are needed
to assess optimization stability.


\begin{table}[!htbp]
    \vspace{-1mm}
    \centering
    \small
    \caption{Inference efficiency on an H20 GPU (3-image)} 
    
    \vspace{-2mm}
    \label{tab:efficiency}
    \begin{tabular}{lcccc}
        \toprule
        Method & KR & \makecell{Lat.\\(ms)$\downarrow$} & \makecell{Mem.\\(MB)$\downarrow$} & \makecell{FLOPs\\(G)$\downarrow$} \\
        \midrule
        Baseline              & 100\% & 714.1 & 1914  & 48216.0 \\
        \midrule
        FlashVLM              & 21\%  & 895.2 & 9174  & 22207.4 \\
        HiPrune               & 21\%  & 789.2 & 15270 & 22207.4 \\
        PDrop                 & 41\%  & 557.1 & \textbf{956}   & 26430.7 \\
        IVC                   & 50\%  & 723.5 & 984   & 39337.8 \\
      Attention & 21\% & 707.7 & 6840 & 35328.4\\
        \midrule
        \textbf{STP-GUI} & 21\% & \textbf{494.5} & 982 & \textbf{20746.4} \\
        \bottomrule
    \end{tabular}
    \vspace{-5.5mm}
\end{table}

\paragraph{Efficiency Analysis.} We benchmark inference on an H20 GPU with 1, 2 and 3 frames (input lengths are 2420 / 4468 / 6517 tokens). Since real-world web-agent trajectories typically involve at least 3 steps, Tab.~\ref{tab:efficiency} reports the 3-frame setting, while 1- and 2-frame results are provided in the Appendix. We compare STP-GUI at KR $\approx 0.21$ against competing methods in terms of latency, memory and FLOPs. At 21\% retention, STP-GUI reduces latency by 31\% and FLOPs by 57\% relative to the dense baseline. STP-GUI achieves the lowest absolute latency (494.5ms) among all evaluated methods. Competing methods like FlashVLM and HiPrune require dynamic LLM attention masks during inference, incurring 9.3$\times$ and 15.5$\times$ higher memory overhead than STP-GUI, respectively (and about 4.8$\times$ and 8.0$\times$ more than the dense baseline). In contrast, STP-GUI operates on ViT features before LLM prefill, maintaining a minimal memory footprint while delivering superior end-to-end speedups.



\vspace{-2mm}

\section{Conclusion}
\vspace{-2mm}
In this paper, we present STP-GUI, a spatially-aware visual token pruning method for Web-GUI agents. By combining query-guided scoring with window-level selection, STP-GUI preserves action-relevant local evidence as contiguous patches under constrained token budgets. On Online-Mind2Web, it retains 93.8\% of full-token performance at 50\% retention and 77.7\% at 21\% retention, while reducing latency by 31\% and FLOPs by 57\%. These results show that lightweight pre-LLM pruning is a practical design for GUI agents operating under aggressive token constraints.
\vspace{-2mm}
\paragraph{Limitations.} Our current study focuses on Web-GUI benchmarks and model scales up to Qwen3.5-9B. Additionally, STP-GUI uses a fixed token budget and predefined non-overlapping window granularity, which may be less suitable for interface elements that span window boundaries. Nevertheless, high-resolution encodings preserve sufficient token signals, and our overlap ablation (Tab.~\ref{tab:overlap}) suggests that coarse locality outweighs strict boundary alignment in practice.


\bibliography{main}

@String(ECCV  = {Eur. Conf. Comput. Vis.})

@String(NeurIPS = {Adv. Neural Inform. Process. Syst.})

@String(ECCV  = {ECCV})

@String(NeurIPS = {NeurIPS})

@inproceedings{attention,
  title={Attention is all you need},
  author={Vaswani, Ashish and Shazeer, Noam and Parmar, Niki and Uszkoreit, Jakob and Jones, Llion and Gomez, Aidan N and Kaiser, {\L}ukasz and Polosukhin, Illia},
  booktitle={NeurIPS},
  year={2017}
}

@article{bai2025qwen3,
  title={Qwen3-vl technical report},
  author={Bai, Shuai and Cai, Yuxuan and Chen, Ruizhe and Chen, Keqin and Chen, Xionghui and Cheng, Zesen and Deng, Lianghao and Ding, Wei and Gao, Chang and Ge, Chunjiang and others},
  journal={arXiv preprint arXiv:2511.21631},
  year={2025}
}

@misc{bai2025qwen25vltechnicalreport,
      title={Qwen2.5-VL Technical Report}, 
      author={Shuai Bai and Keqin Chen and Xuejing Liu and Jialin Wang and Wenbin Ge and Sibo Song and Kai Dang and Peng Wang and Shijie Wang and Jun Tang and Humen Zhong and Yuanzhi Zhu and Mingkun Yang and Zhaohai Li and Jianqiang Wan and Pengfei Wang and Wei Ding and Zheren Fu and Yiheng Xu and Jiabo Ye and Xi Zhang and Tianbao Xie and Zesen Cheng and Hang Zhang and Zhibo Yang and Haiyang Xu and Junyang Lin},
      year={2025},
      eprint={2502.13923},
      archivePrefix={arXiv},
      primaryClass={cs.CV},
      url={https://arxiv.org/abs/2502.13923}, 
}

@misc{qin2025uitarspioneeringautomatedgui,
      title={UI-TARS: Pioneering Automated GUI Interaction with Native Agents}, 
      author={Yujia Qin and Yining Ye and Junjie Fang and Haoming Wang and Shihao Liang and Shizuo Tian and Junda Zhang and Jiahao Li and Yunxin Li and Shijue Huang and Wanjun Zhong and Kuanye Li and Jiale Yang and Yu Miao and Woyu Lin and Longxiang Liu and Xu Jiang and Qianli Ma and Jingyu Li and Xiaojun Xiao and Kai Cai and Chuang Li and Yaowei Zheng and Chaolin Jin and Chen Li and Xiao Zhou and Minchao Wang and Haoli Chen and Zhaojian Li and Haihua Yang and Haifeng Liu and Feng Lin and Tao Peng and Xin Liu and Guang Shi},
      year={2025},
      eprint={2501.12326},
      archivePrefix={arXiv},
      primaryClass={cs.AI},
      url={https://arxiv.org/abs/2501.12326}, 
}

@article{wang2025ui,
  title={Ui-tars-2 technical report: Advancing gui agent with multi-turn reinforcement learning},
  author={Wang, Haoming and Zou, Haoyang and Song, Huatong and Feng, Jiazhan and Fang, Junjie and Lu, Junting and Liu, Longxiang and Luo, Qinyu and Liang, Shihao and Huang, Shijue and others},
  journal={arXiv preprint arXiv:2509.02544},
  year={2025}
}

@article{cai2025flashvlm,
  title={Flashvlm: Text-guided visual token selection for large multimodal models},
  author={Cai, Kaitong and Zhang, Jusheng and Yang, Jing and Fan, Yijia and Xie, Pengtao and Wang, Jian and Wang, Keze},
  journal={arXiv preprint arXiv:2512.20561},
  year={2025}
}

@article{wang2024cogvlm,
  title={Cogvlm: Visual expert for pretrained language models},
  author={Wang, Weihan and Lv, Qingsong and Yu, Wenmeng and Hong, Wenyi and Qi, Ji and Wang, Yan and Ji, Junhui and Yang, Zhuoyi and Zhao, Lei and XiXuan, Song and others},
  journal={Advances in Neural Information Processing Systems},
  volume={37},
  pages={121475--121499},
  year={2024}
}

@article{wang2024qwen2,
  title={Qwen2-vl: Enhancing vision-language model's perception of the world at any resolution},
  author={Wang, Peng and Bai, Shuai and Tan, Sinan and Wang, Shijie and Fan, Zhihao and Bai, Jinze and Chen, Keqin and Liu, Xuejing and Wang, Jialin and Ge, Wenbin and others},
  journal={arXiv preprint arXiv:2409.12191},
  year={2024}
}

@inproceedings{hong2024cogagent,
  title={Cogagent: A visual language model for gui agents},
  author={Hong, Wenyi and Wang, Weihan and Lv, Qingsong and Xu, Jiazheng and Yu, Wenmeng and Ji, Junhui and Wang, Yan and Wang, Zihan and Dong, Yuxiao and Ding, Ming and others},
  booktitle={Proceedings of the IEEE/CVF conference on computer vision and pattern recognition},
  pages={14281--14290},
  year={2024}
}

@inproceedings{
liu2026hiprunehierarchicalattentionefficient,
  title={HiPrune: Hierarchical Attention for Efficient Token Pruning in Vision-Language Models},
  author={Jizhihui, Liu and Guangdao, Zhu and Feiyi, Du and Niu, Lian and Jun, Li and Bin, Chen and Weili, Guan and Yaowei, Wang},
  booktitle={The 64th Annual Meeting of the Association for Computational Linguistics},
  year={2026}
}

@article{xing2024pyramiddrop,
  title={Pyramiddrop: Accelerating your large vision-language models via pyramid visual redundancy reduction},
  author={Xing, Long and Huang, Qidong and Dong, Xiaoyi and Lu, Jiajie and Zhang, Pan and Zang, Yuhang and Cao, Yuhang and He, Conghui and Wang, Jiaqi and Wu, Feng and others},
  journal={arXiv preprint arXiv:2410.17247},
  year={2024}
}

@inproceedings{chen2025guicourse,
  title={Guicourse: From general vision language model to versatile gui agent},
  author={Chen, Wentong and Cui, Junbo and Hu, Jinyi and Qin, Yujia and Fang, Junjie and Zhao, Yue and Wang, Chongyi and Liu, Jun and Chen, Guirong and Huo, Yupeng and others},
  booktitle={Proceedings of the 63rd Annual Meeting of the Association for Computational Linguistics (Volume 1: Long Papers)},
  pages={21936--21959},
  year={2025}
}

@misc{rawles2023androidwildlargescaledataset,
      title={Android in the Wild: A Large-Scale Dataset for Android Device Control}, 
      author={Christopher Rawles and Alice Li and Daniel Rodriguez and Oriana Riva and Timothy Lillicrap},
      year={2023},
      eprint={2307.10088},
      archivePrefix={arXiv},
      primaryClass={cs.LG},
      url={https://arxiv.org/abs/2307.10088}, 
}

@article{tang2025magicgui,
  title={Magicgui: A foundational mobile gui agent with scalable data pipeline and reinforcement fine-tuning},
  author={Tang, Liujian and Dong, Shaokang and Huang, Yijia and Xiang, Minqi and Ruan, Hongtao and Wang, Bin and Li, Shuo and Xi, Zhiheng and Cao, Zhihui and Pang, Hailiang and others},
  journal={arXiv preprint arXiv:2508.03700},
  year={2025}
}

@article{bolya2022token,
  title={Token merging: Your vit but faster},
  author={Bolya, Daniel and Fu, Cheng-Yang and Dai, Xiaoliang and Zhang, Peizhao and Feichtenhofer, Christoph and Hoffman, Judy},
  journal={arXiv preprint arXiv:2210.09461},
  year={2022}
}

@inproceedings{chen2024image,
  title={An image is worth 1/2 tokens after layer 2: Plug-and-play inference acceleration for large vision-language models},
  author={Chen, Liang and Zhao, Haozhe and Liu, Tianyu and Bai, Shuai and Lin, Junyang and Zhou, Chang and Chang, Baobao},
  booktitle={European Conference on Computer Vision},
  pages={19--35},
  year={2024},
  organization={Springer}
}

@inproceedings{alvar2025divprune,
  title={Divprune: Diversity-based visual token pruning for large multimodal models},
  author={Alvar, Saeed Ranjbar and Singh, Gursimran and Akbari, Mohammad and Zhang, Yong},
  booktitle={Proceedings of the Computer Vision and Pattern Recognition Conference},
  pages={9392--9401},
  year={2025}
}

@inproceedings{wen2025stop,
  title={Stop Looking for “Important Tokens” in Multimodal Language Models: Duplication Matters More},
  author={Wen, Zichen and Gao, Yifeng and Wang, Shaobo and Zhang, Junyuan and Zhang, Qintong and Li, Weijia and He, Conghui and Zhang, Linfeng},
  booktitle={Proceedings of the 2025 Conference on Empirical Methods in Natural Language Processing},
  pages={9972--9991},
  year={2025}
}

@inproceedings{
shen2025fastvid,
title={Fast{VID}: Dynamic Density Pruning for Fast Video Large Language Models},
author={Leqi Shen and Guoqiang Gong and Tao He and Yifeng Zhang and pengzhang liu and Sicheng Zhao and Guiguang Ding},
booktitle={The Thirty-ninth Annual Conference on Neural Information Processing Systems},
year={2025},
url={https://openreview.net/forum?id=2xS4VtpApy}
}

@article{zhang2024sparsevlm,
  title={Sparsevlm: Visual token sparsification for efficient vision-language model inference},
  author={Zhang, Yuan and Fan, Chun-Kai and Ma, Junpeng and Zheng, Wenzhao and Huang, Tao and Cheng, Kuan and Gudovskiy, Denis and Okuno, Tomoyuki and Nakata, Yohei and Keutzer, Kurt and others},
  journal={arXiv preprint arXiv:2410.04417},
  year={2024}
}

@article{liu2024multi,
  title={Multi-stage vision token dropping: Towards efficient multimodal large language model},
  author={Liu, Ting and Shi, Liangtao and Hong, Richang and Hu, Yue and Yin, Quanjun and Zhang, Linfeng},
  journal={arXiv preprint arXiv:2411.10803},
  year={2024}
}

@inproceedings{ye2025atp,
  title={Atp-llava: Adaptive token pruning for large vision language models},
  author={Ye, Xubing and Gan, Yukang and Ge, Yixiao and Zhang, Xiao-Ping and Tang, Yansong},
  booktitle={Proceedings of the IEEE/CVF Conference on Computer Vision and Pattern Recognition},
  pages={24972--24982},
  year={2025}
}

@article{hurst2024gpt-4o,
  title={Gpt-4o system card},
  author={Hurst, Aaron and Lerer, Adam and Goucher, Adam P and Perelman, Adam and Ramesh, Aditya and Clark, Aidan and Ostrow, AJ and Welihinda, Akila and Hayes, Alan and Radford, Alec and others},
  journal={arXiv preprint arXiv:2410.21276},
  year={2024}
}

@article{xia2024agentless,
  title={Agentless: Demystifying llm-based software engineering agents},
  author={Xia, Chunqiu Steven and Deng, Yinlin and Dunn, Soren and Zhang, Lingming},
  journal={arXiv preprint arXiv:2407.01489},
  year={2024}
}

@inproceedings{gao2024enhancing,
  title={Enhancing vision-language pre-training with rich supervisions},
  author={Gao, Yuan and Shi, Kunyu and Zhu, Pengkai and Belval, Edouard and Nuriel, Oren and Appalaraju, Srikar and Ghadar, Shabnam and Tu, Zhuowen and Mahadevan, Vijay and Soatto, Stefano},
  booktitle={Proceedings of the IEEE/CVF conference on computer vision and pattern recognition},
  pages={13480--13491},
  year={2024}
}

@inproceedings{shang2025llava,
  title={Llava-prumerge: Adaptive token reduction for efficient large multimodal models},
  author={Shang, Yuzhang and Cai, Mu and Xu, Bingxin and Lee, Yong Jae and Yan, Yan},
  booktitle={Proceedings of the IEEE/CVF International Conference on Computer Vision},
  pages={22857--22867},
  year={2025}
}

@article{xie2024osworld,
  title={Osworld: Benchmarking multimodal agents for open-ended tasks in real computer environments},
  author={Xie, Tianbao and Zhang, Danyang and Chen, Jixuan and Li, Xiaochuan and Zhao, Siheng and Cao, Ruisheng and Hua, Toh J and Cheng, Zhoujun and Shin, Dongchan and Lei, Fangyu and others},
  journal={Advances in Neural Information Processing Systems},
  volume={37},
  pages={52040--52094},
  year={2024}
}

@inproceedings{zheng2024llamafactory,
  title={Llamafactory: Unified efficient fine-tuning of 100+ language models},
  author={Zheng, Yaowei and Zhang, Richong and Zhang, Junhao and Ye, Yanhan and Luo, Zheyan},
  booktitle={Proceedings of the 62nd annual meeting of the association for computational linguistics (volume 3: system demonstrations)},
  pages={400--410},
  year={2024}
}

@article{xue2025illusion,
  title={An illusion of progress? assessing the current state of web agents},
  author={Xue, Tianci and Qi, Weijian and Shi, Tianneng and Song, Chan Hee and Gou, Boyu and Song, Dawn and Sun, Huan and Su, Yu},
  journal={arXiv preprint arXiv:2504.01382},
  year={2025}
}

@misc{sun2026ivcprunerevealingimplicitvisual,
      title={IVC-Prune: Revealing the Implicit Visual Coordinates in LVLMs for Vision Token Pruning}, 
      author={Zhichao Sun and Yidong Ma and Gang Liu and Yibo Chen and Xu Tang and Yao Hu and Yongchao Xu},
      year={2026},
      eprint={2602.03060},
      archivePrefix={arXiv},
}

@inproceedings{cha2024honeybee,
  title={Honeybee: Locality-enhanced projector for multimodal llm},
  author={Cha, Junbum and Kang, Wooyoung and Mun, Jonghwan and Roh, Byungseok},
  booktitle={Proceedings of the IEEE/CVF Conference on Computer Vision and Pattern Recognition},
  pages={13817--13827},
  year={2024}
}

@article{fu2025mano,
  title={Mano Technical Report},
  author={Fu, Tianyu and Su, Anyang and Zhao, Chenxu and Wang, Hanning and Wu, Minghui and Yu, Zhe and Hu, Fei and Shi, Mingjia and Dong, Wei and Wang, Jiayao and others},
  journal={arXiv preprint arXiv:2509.17336},
  year={2025}
}

@article{jeddi2025similarity,
  title={Similarity-aware token pruning: Your vlm but faster},
  author={Jeddi, Ahmadreza and Baghbanzadeh, Negin and Dolatabadi, Elham and Taati, Babak},
  journal={arXiv preprint arXiv:2503.11549},
  year={2025}
}

@inproceedings{yang2025visionzip,
  title={Visionzip: Longer is better but not necessary in vision language models},
  author={Yang, Senqiao and Chen, Yukang and Tian, Zhuotao and Wang, Chengyao and Li, Jingyao and Yu, Bei and Jia, Jiaya},
  booktitle={Proceedings of the IEEE/CVF Conference on Computer Vision and Pattern Recognition},
  pages={19792--19802},
  year={2025}
}

@article{bai2024digirl,
  title={Digirl: Training in-the-wild device-control agents with autonomous reinforcement learning},
  author={Bai, Hao and Zhou, Yifei and Cemri, Mert and Pan, Jiayi and Suhr, Alane and Levine, Sergey and Kumar, Aviral},
  journal={Advances in Neural Information Processing Systems},
  volume={37},
  pages={12461--12495},
  year={2024}
}

@inproceedings{ho2022yoro,
  title={{YORO}: Lightweight End to End Visual Grounding},
  author={Ho, Chih-Hui and Appalaraju, Srikar and Jasani, Bhavan and Manmatha, R. and Vasconcelos, Nuno},
  booktitle={ECCV Workshop on International Challenge on Compositional and Multimodal Perception},
  year={2022}
}

@inproceedings{feng2025towards,
  title={Towards Efficient Online Tuning of VLM Agents via Counterfactual Soft Reinforcement Learning},
  author={Feng, Lang and Tan, Weihao and Lyu, Zhiyi and Zheng, Longtao and Xu, Haiyang and Yan, Ming and Huang, Fei and An, Bo},
  booktitle={International Conference on Machine Learning},
  pages={16884--16903},
  year={2025},
  organization={PMLR}
}

@inproceedings{park2025rvlm,
  title={{R-VLM}: Region-Aware Vision Language Model for Precise {GUI} Grounding},
  author={Park, Joonhyung and Tang, Peng and Das, Sagnik and Appalaraju, Srikar and Singh, Kunwar Yashraj and Manmatha, R. and Ghadar, Shabnam},
  booktitle={Proceedings of the 63rd Annual Meeting of the Association for Computational Linguistics},
  year={2025}
}

@misc{qwen3.5,
    title  = {{Qwen3.5}: Towards Native Multimodal Agents},
    author = {{Qwen Team}},
    year   = {2026},
    month  = {February},
    url    = {https://qwen.ai/blog?id=qwen3.5}
}

\clearpage
\appendix
\section{Appendix}
\subsection{More Implementation Details}
For comparison methods, we re-implement each method~\cite{xing2024pyramiddrop,liu2026hiprunehierarchicalattentionefficient,sun2026ivcprunerevealingimplicitvisual,cai2025flashvlm} on top of Qwen3-VL using their official codebases. To ensure a fair comparison, we evaluate both their original training-free configurations and fine-tuned variants adapted on the same SFT data. Layer-sensitive baselines are evaluated using their best-performing configurations. Detailed sensitivity analyses are provided in appendix. Because Online-Mind2Web is conducted on live websites, we further assess evaluation stability by repeating all experiments 3 times under the same protocol and report the mean success rate in the main results. Across runs, the difference between the highest and lowest success rates never exceeds 5 successful tasks out of 300 (0.017 absolute SR). Since the performance gap between STP-GUI and the strongest competing baseline is larger than this fluctuation range, the main conclusions are unlikely to be explained by run-to-run evaluation noise alone. We use the action space defined in the agent prompt, which includes discrete action types with optional text or spatial arguments, such as clicking, hotkey invocation, scrolling, and termination.

\subsection{Layer Selection Sensitivity}
\label{sec:appendix_layer_selection}

For layer-dependent methods (IVC and Attention), we evaluate three layer configurations: shallow (layer 4), middle (layer 8), and deep (layer 16). Table~\ref{tab:layer_sensitivity} reports SR across configurations for both methods. Deep layers (layer 16) consistently yield the best results, consistent with the view that task-relevant localization signals emerge in later transformer layers after the model has processed both visual context and instruction semantics. We use the best-performing configuration for each method in the main experiments.

\begin{table}[h]
\centering
\caption{SR across layer configurations on Qwen3-VL-2B. Best result per group in \textbf{bold}.}
\label{tab:layer_sensitivity}
\setlength{\tabcolsep}{14pt}
\begin{tabular}{llcc}
    \toprule
    Method & KR & Layer & SR ($\uparrow$) \\
    \midrule
    \multirow{3}{*}{IVC} & \multirow{3}{*}{0.50}
        & 4  & 0.130 \\
        & & 8  & 0.177 \\
        & & 16 & \textbf{0.217} \\
    \midrule
    \multirow{12}{*}{Attention}
        & \multirow{3}{*}{0.10} & 4  & 0.060 \\
        &                       & 8  & 0.100 \\
        &                       & 16 & \textbf{0.130} \\
        \cmidrule{2-4}
        & \multirow{3}{*}{0.21} & 4  & 0.090 \\
        &                       & 8  & 0.147 \\
        &                       & 16 & \textbf{0.220} \\
        \cmidrule{2-4}
        & \multirow{3}{*}{0.28} & 4  & 0.127 \\
        &                       & 8  & 0.170 \\
        &                       & 16 & \textbf{0.250} \\
        \cmidrule{2-4}
        & \multirow{3}{*}{0.49} & 4  & 0.130 \\
        &                       & 8  & 0.173 \\
        &                       & 16 & \textbf{0.280} \\
    \bottomrule
\end{tabular}
\end{table}

\subsection{Training Cost}
Table~\ref{tab:training_cost} reports the tuning cost of STP-GUI and the adapted training-based baselines under the same Qwen3-VL-2B setting. All adapted methods are trained with their official implementations on the same SFT data used to train STP-GUI.
The adapted methods require comparable or larger optimization budgets than STP-GUI. For example, HiPrune requires 10.3--14.1 hours and Attention requires 9.7--11.0 hours, while STP-GUI takes 4.8 hours when trained with ground-truth rewards only and 10.6 hours with the combined reward. Unlike the baselines, which update a subset of VLM parameters during training, STP-GUI only optimizes a lightweight 2M-parameter selector while keeping the VLM backbone fully frozen. When ground-truth action supervision is available, STP-GUI can be trained without the additional cost of offline full-model rollouts, making the GT-only variant a practical option for budget-constrained settings.

\begin{table}[h]
\centering
\caption{Tuning cost on Qwen3-VL-2B (61,427 samples, 8$\times$A100 GPUs). All times are wall-clock hours. STP-GUI optimizes only the 2M-parameter selector with the VLM backbone frozen.}
\label{tab:training_cost}
\setlength{\tabcolsep}{8pt}
\begin{tabular}{lcc}
    \toprule
    Method & KR & \makecell{Training \\Time (h)} \\
    \midrule
    \multirow{2}{*}{FlashVLM}  & 0.21 & 5.3 \\
    & 0.50 & 6.2 \\
    \multirow{2}{*}{HiPrune}   & 0.21 & 10.3 \\
     & 0.50 & 14.1 \\
    PDrop     & 0.41 & 2.1 \\
    IVC       & 0.50 & 7.0 \\
    \multirow{2}{*}{Attention} & 0.21 & 9.7 \\
     & 0.50 & 11.0 \\
    \midrule
    STP-GUI (GT reward only)    & 0.21 & 4.8 \\
    STP-GUI (GT + full reward)  & 0.21 & 10.6 \\
    \bottomrule
\end{tabular}
\end{table}

\subsection{8-Connectivity Connected-Component Analysis}
\label{sec:appendix_8connectivity}

The main paper reports connected-component statistics under 4-connectivity to quantify spatial fragmentation. We additionally report the same statistics under 8-connectivity. The 4 metrics---number of connected components ($N_{\mathrm{cc}}$, $\downarrow$), largest component ratio ($R_{\mathrm{largest}}$, $\uparrow$), mean component area ($A_{\mathrm{mean}}$, $\uparrow$), and component-size entropy ($H_{\mathrm{cc}}$, $\downarrow$)---are identical to those in the main paper; the only difference is the adjacency rule.

\begin{table}[h]
\centering
\caption{Connected-component statistics under 8-connectivity on Online-Mind2Web (Qwen3-VL-2B, keep ratio 0.21). Mean$\pm$std.\ over 300 trajectories.}
\scriptsize 
\label{tab:8connectivity}
\setlength{\tabcolsep}{3pt}
\begin{tabular}{lcccc}
    \toprule
    Method & $N_{\mathrm{cc}}$$\downarrow$ & $R_{\mathrm{largest}}$$\uparrow$ & $A_{\mathrm{mean}}$$\uparrow$ & $H_{\mathrm{cc}}$$\downarrow$ \\
    \midrule
    STP-GUI   & \textbf{5.29$\pm$2.16}  & \textbf{0.794$\pm$0.180} & \textbf{133.4$\pm$75.4} & \textbf{0.665$\pm$0.448} \\
    FlashVLM  & 82.82$\pm$12.49 & 0.215$\pm$0.077 & 4.91$\pm$0.86   & 3.383$\pm$0.399 \\
    Attention & 71.45$\pm$11.63 & 0.311$\pm$0.127 & 5.86$\pm$1.03   & 2.894$\pm$0.439 \\
    HiPrune   & 97.88$\pm$9.41  & 0.104$\pm$0.038 & 4.08$\pm$0.39   & 3.961$\pm$0.171 \\
    \bottomrule
\end{tabular}
\end{table}

The trends are similar to those under 4-connectivity. Under 8-connectivity, STP-GUI produces 5.29 connected components on average, with 79.4\% of retained tokens concentrated in the largest component. In contrast, FlashVLM, Attention, and HiPrune produce 71--98 components, with the largest component accounting for only 10--31\% of retained tokens. The component-size entropy of STP-GUI (0.665) is also substantially lower than that of the baselines (2.89--3.96), confirming that the retained tokens are concentrated in a small number of coherent regions rather than scattered across many isolated patches. These results indicate that the spatial coherence advantage of STP-GUI is robust to the choice of connectivity rule.

\subsection{Results on Qwen3.5-9B}
\label{sec:appendix_qwen35}

To examine whether STP-GUI generalizes to larger and more recent VLM backbones, we additionally evaluate it on Qwen3.5-9B. The backbone is first fine-tuned on the GUI training data following the same protocol as the Qwen3-VL models, after which the STP-GUI selector is trained on top of the frozen backbone. We evaluate on AitW-General and Online-Mind2Web.
\begin{table}[h]
\centering
\caption{Results on Qwen3.5-9B. Baseline uses all visual tokens; ACC denotes the ratio relative to the full-token baseline.}

\label{tab:qwen35_9b}
\setlength{\tabcolsep}{10pt}
\begin{tabular}{lcccc}
    \toprule
    Dataset & KR & SR ($\uparrow$) & ACC ($\uparrow$) \\
    \midrule
    \multirow{3}{*}{Mind2Web} & 100\% & 0.42 & 100\% \\
    & 21\% & 0.319 & 75.95\% \\
    & 50\% & 0.387 & 92.14\% \\
     \cmidrule{2-4}
    \multirow{2}{*}{AitW-General}
        & 100\% & 0.8356 & 100\% \\
        & 21\% & 0.7083 & 84.76\% \\
        & 50\% & 0.7546 & 90.31\% \\
    \bottomrule
\end{tabular}
\end{table}

At 50\% retention, STP-GUI maintains 90.31\% of the full-token performance on AitW-General (0.7546 vs.\ 0.8356) and 92.14\% on Online-Mind2Web (0.387 vs.\ 0.42). The performance retention rate is comparable to that observed on Qwen3-VL-2B (93.8\%) and Qwen3-VL-4B (95.0\%), suggesting that the effectiveness of spatially coherent token selection is not tied to a specific model scale or architecture generation.

\subsection{Efficiency Analysis: 1-Frame and 2-Frame Settings}
\label{sec:appendix_efficiency_1_2frame}

Table~\ref{tab:efficiency_1frame} and Table~\ref{tab:efficiency_2frame} report inference efficiency for the 1-frame (2420 tokens) and 2-frame (4468 tokens) scenarios on an H20 GPU, complementing the 3-frame results in Tab.~\ref{tab:efficiency} of the main paper. All settings use keep ratio $\approx 0.21$ with STP-GUI at $W=4$.

\begin{table}[h]
\caption{Inference efficiency at keep ratio $\approx$ 0.21 on an H20 GPU (1-image scenario, 2420 tokens). STP-GUI uses $W=4$. Lower is better for all metrics.}
\label{tab:efficiency_1frame}
\begin{tabular}{lcccc}
    \toprule
    Method & KR & \makecell{Lat.\\(ms)$\downarrow$} & \makecell{Mem.\\(MB)$\downarrow$} & \makecell{FLOPs\\(G)$\downarrow$} \\
    \midrule
    Baseline              & 100\% & 244.6 & 712  & 15016.8 \\
    \midrule
    FlashVLM              & 21\%  & 298.5 & 4759 &  8196.3 \\
    HiPrune               & 21\%  & 286.6 & 6791 &  8196.3 \\
    PDrop                 & 41\%  & 204.1 & \textbf{318} &  9302.0 \\
    IVC                   & 50\%  & 236.4 & 388  & 12601.9 \\
    Attention             & 21\%  & 232.2 & 1005 & 11513.3 \\
    \midrule
    \textbf{STP-GUI}         & 21\%  & \textbf{189.3} & 328 & \textbf{8110.6} \\
    \bottomrule
\end{tabular}
\end{table}

\begin{table}[]
\centering
\small
\caption{Inference efficiency at keep ratio $\approx$ 0.21 on an H20 GPU (2-image scenario, 4468 tokens). STP-GUI uses $W=4$. Lower is better for all metrics.}
\label{tab:efficiency_2frame}
\begin{tabular}{lcccc}
    \toprule
    Method & KR & \makecell{Lat.\\(ms)$\downarrow$} & \makecell{Mem.\\(MB)$\downarrow$} & \makecell{FLOPs\\(G)$\downarrow$} \\
    \midrule
    Baseline              & 100\% & 476.8 & 1312  & 30649.2 \\
    \midrule
    FlashVLM              & 21\%  & 576.0 & 6966  & 15160.0 \\
    HiPrune               & 21\%  & 538.0 & 11031 & 15160.0 \\
    PDrop                 & 41\%  & 379.1 & \textbf{636} & 17668.6 \\
    IVC                   & 50\%  & 472.1 & 683   & 25294.4 \\
    Attention             & 21\%  & 458.5 & 3275  & 22850.2 \\
    \midrule
    \textbf{STP-GUI}         & 21\%  & \textbf{341.1} & 656 & \textbf{14256.5} \\
    \bottomrule
\end{tabular}
\end{table}

\begin{figure}[!htbp]
\vspace{-3mm}
    \centering
    \includegraphics[width=0.85\linewidth]{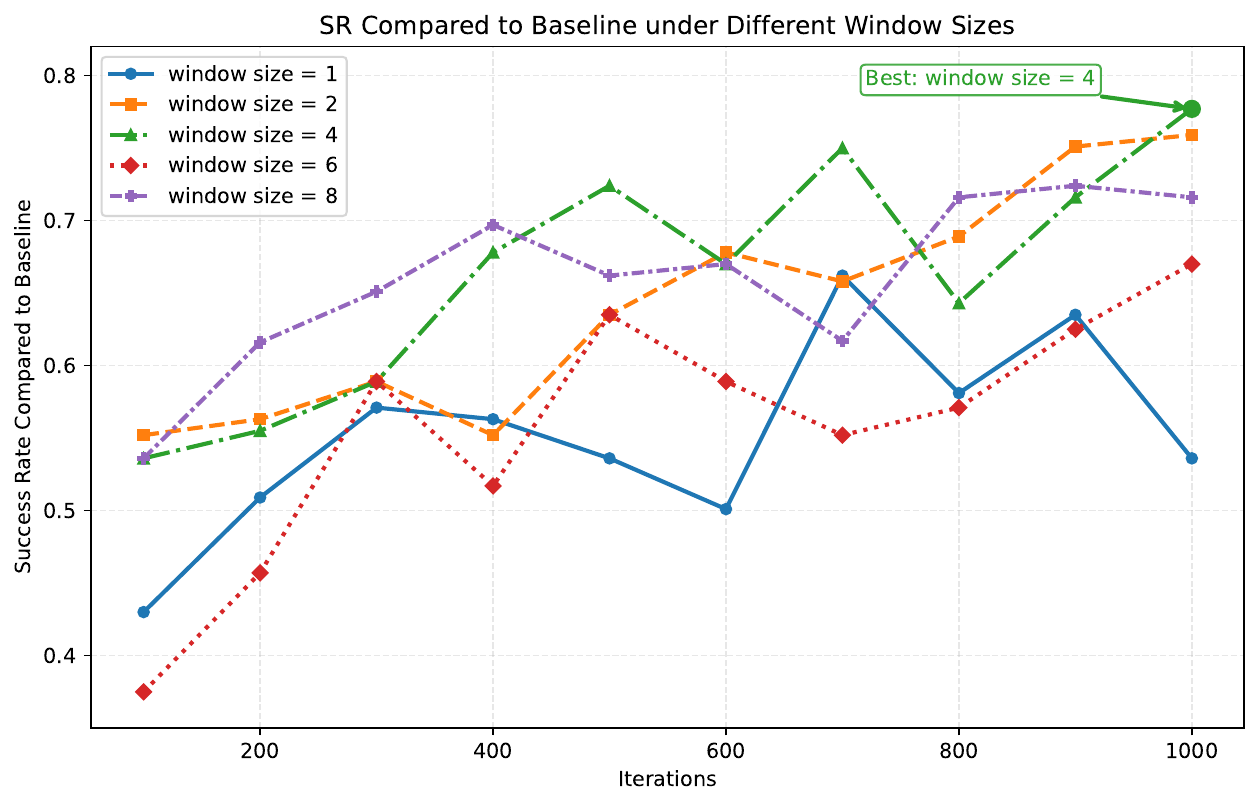}
    \vspace{-4mm}
    \caption{SR vs.\ training iterations for different window sizes on Online-Mind2Web. $W=4$ achieves best-performing window granularity under our GUI settings\,(0.290).}
    \label{fig:window_size_ablation}
    \vspace{-4mm}
\end{figure}

\subsection{Effect of window size.}
Window size $W$ controls the spatial granularity of token selection. We study its effect by evaluating $W \in \{1, 2, 4, 6, 8\}$ across training iterations on Online-Mind2Web (Qwen3-VL-2B, KR = 0.21), with all other hyperparameters fixed. Fig.~\ref{fig:window_size_ablation} reports SR at each checkpoint. When $W=1$, the selector degenerates to token-wise selection, exhibiting high variance throughout training and the lowest final SR. To better understand the effect of selection granularity, we further visualize the retained visual tokens of several pruning methods in appendix. Under the similar token budget, these methods tend to keep visually scattered patches, while STP-GUI preserves more spatially contiguous regions around the target element. The difference helps explain the window-size ablation results: when \(W{=}1\), the selector degenerates to token-wise selection and is more likely to fragment the local evidence needed for grounding, whereas moderately sized windows better preserve action-relevant local structure. The training curves cross across checkpoints, while
$W=4$ achieves the highest reported final SR among
the tested settings.
Small windows impose weaker local grouping, whereas
large windows may allocate more tokens to irrelevant
regions.
This tradeoff motivates the default window size,
although the experiment does not separate
target-boundary coverage from optimization effects.

\subsection{Effect of Reward Composition.}

\begin{table}[!htbp]
    \vspace{-2mm}
    \centering
    \small
    \setlength{\tabcolsep}{17pt}
    \caption{Ablation on reward composition $(\lambda_{\mathrm{gt}},\, \lambda_{\mathrm{full}})$. Qwen3-VL-2B with $W=4$ and ratio = 0.21.}
    \label{tab:reward_ablation2}
    \begin{tabular}{llcc}
        \toprule
        $\lambda_{\mathrm{gt}}$ & $\lambda_{\mathrm{full}}$ & SR($\uparrow$) &ACC($\uparrow$) \\
        \midrule
        \multicolumn{2}{c}{Baseline}  & 0.373 &100\%\\
        \midrule
        1.0 & 0.0 & 0.280 & 75.1\% \\  
        0.0 & 1.0 & 0.283 & 75.9\% \\  
        0.7 & 0.3 & \textbf{0.290} & \textbf{77.7\%} \\ 
        0.3 & 0.7 & 0.287 & 76.9\% \\  
        0.5 & 0.5 & 0.256 & 68.6\% \\  
        \bottomrule
    \end{tabular}
    \vspace{-5mm}
\end{table}

Tab.~\ref{tab:reward_ablation2} ablates the weighting between the ground-truth reward $r_{\mathrm{gt}}$ and the full-context reward $r_{\mathrm{full}}$ in Eq.~\ref{eq:reward}. Using $r_{\mathrm{gt}}$ alone (0.280) or $r_{\mathrm{full}}$ alone (0.283) already yields competitive results, confirming that the selector can be trained without ground-truth supervision. Combining both signals ($\lambda_{\mathrm{gt}}=0.7$, $\lambda_{\mathrm{full}}=0.3$) achieves the best result (0.290), indicating that the two rewards provide complementary learning signal. More detailed comparison in Appendix.

\begin{algorithm}[h!]
    \caption{SQTS Training Procedure}
    \label{alg:sqts_alg_abs}

    \SetKwInOut{Input}{Input}
    \SetKwInOut{Output}{Output}

    \Input{Visual tokens $\mathbf{V}$; query $\mathbf{T}$; window size $W$; keep ratio $\beta$;  selector parameters $\theta$; moving scale $\gamma$.}
    \Output{Updated selector parameters $\theta$.}
   
    $ b_{ema} \leftarrow 0$
   
    $\{s_{w_k}\} \leftarrow \mathrm{\phi}_{W \times W}(f_\theta(\mathbf{V}, \mathbf{T}))$ 
   
    $\mathbf{w} \sim \mathrm{PlackettLuce}(\{s_{w_k}\},\; \lfloor\beta L_v / W^2\rfloor)$ 
    \tcp*{Frozen VLM rollout}
    $\hat{a} \leftarrow \mathrm{VLM}(\mathbf{V}[\mathbf{w}], \mathbf{T})$
    
    $a \leftarrow  \text{parse\_action}(\hat{a}) $
    $r \leftarrow \lambda_{\mathrm{gt}} r_{\mathrm{gt}}(\hat{a}) + \lambda_{\mathrm{full}} r_{\mathrm{full}}(\hat{a})$\;
    $A \leftarrow r - b_{ema}(a) $;

    \For{$k$ optimization steps}{
        $\{s_{w_k}'\} \leftarrow \mathrm{\phi}_{W \times W}(f_\theta(\mathbf{V}, \mathbf{T}))$\;
        $\rho \leftarrow \pi_\theta(\mathbf{w} \mid \{s_{w_k}'\}) / \pi_{\theta_{\mathrm{old}}}(\mathbf{w} \mid \{s_{w_k}\})$\;
        $\mathcal{L}_{\mathrm{clip}} \leftarrow -\mathbb{E}[\min(\rho A,\; \mathrm{clip}(\rho, 1{-}\epsilon, 1{+}\epsilon) A)]$\;
        $\theta \leftarrow \theta - \eta \nabla_\theta \left( \mathcal{L}_{\mathrm{clip}} - \beta_H \mathcal{H}(\pi_\theta \mid \{s_{w_k}'\}) \right)$
    }
    $b_{ema}(a) \leftarrow \gamma b_{ema}(a) + (1-\gamma) r$;
\end{algorithm}

\subsection{Step-wise Visualization of Fragmentation Patterns}
\label{app:fragmentation_patterns}

\begin{figure*}[t]
    \centering
\includegraphics[width=0.93\textwidth]{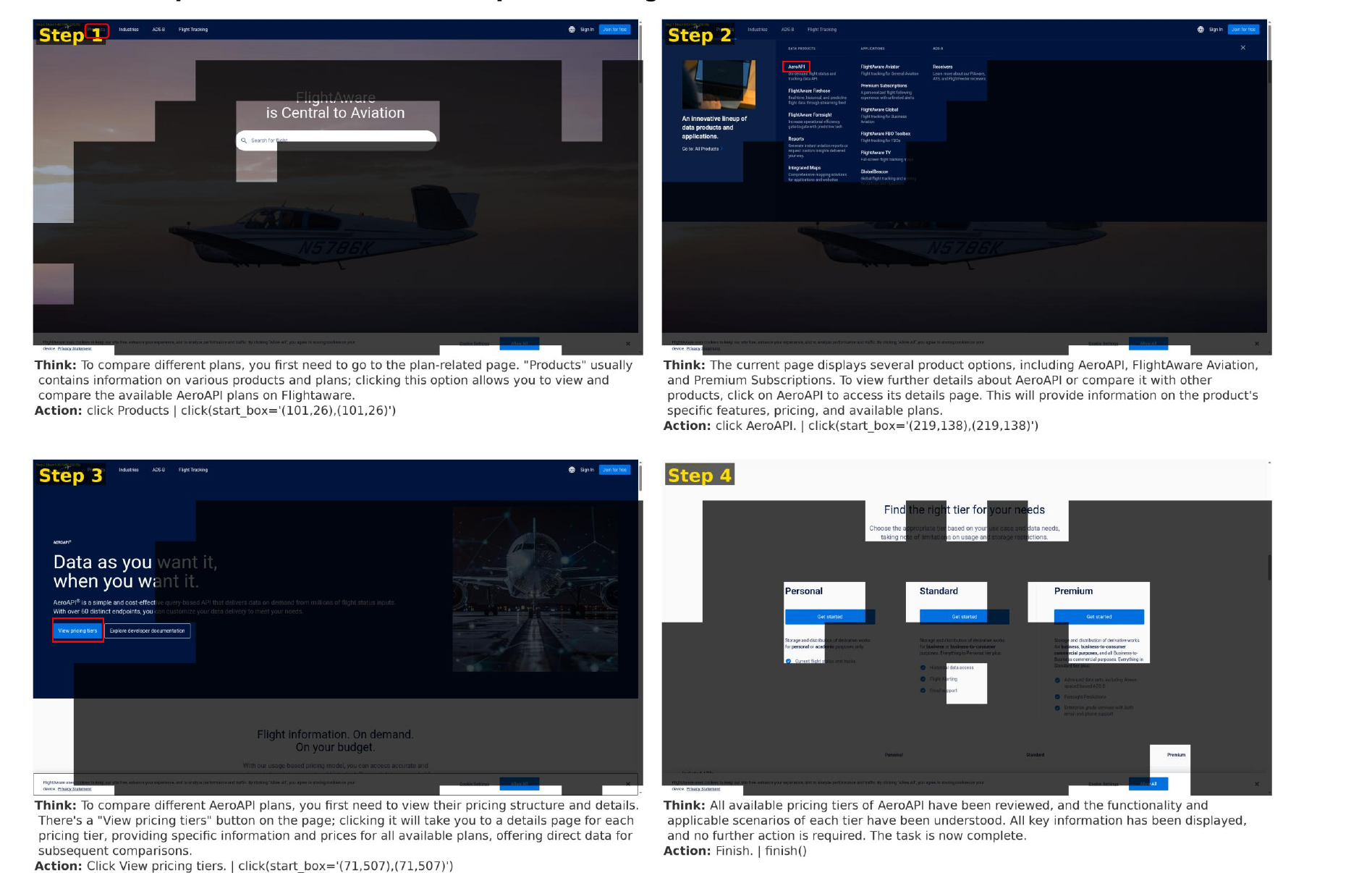}
    \caption{Case-study visualization of STP-GUI token selection across four steps of a successful Online-Mind2Web task (``Compare AeroAPI's available plans''). Shaded and highlighted areas indicate pruned and retained tokens, respectively, while red boxes mark manually annotated step-level targets. Although such annotations are not available benchmark-wide, this representative example shows that STP-GUI preserves spatially coherent regions around key UI elements (e.g., menus, pricing tables) despite significant token reduction.}
    \label{fig:vis_example}
    \vspace{-4mm}
\end{figure*}

Fig.~\ref{fig:vis_example} shows a qualitative visualization of STP-GUI's token selection across 4 sequential steps of a representative Online-Mind2Web task. Because Online-Mind2Web and AitW do not provide dense per-step target annotations for complete trajectories, and successful interaction paths are not unique, we do not treat such examples as a dataset-level metric. Instead, we manually curate several successful Online-Mind2Web episodes as case studies, annotate the step-level ground-truth targets for those representative steps, and compare them with the retained token regions. In these cases, the selector concentrates the token budget on spatially coherent regions (navigation menus, product cards, and pricing tables), while discarding large uniform backgrounds. Token-wise pruning risks fragmenting contiguous GUI elements across the retention boundary; window-level selection instead preserves the local structure that accurate action grounding requires.

\begin{figure*}[t]
    \centering
    \includegraphics[width=\textwidth]{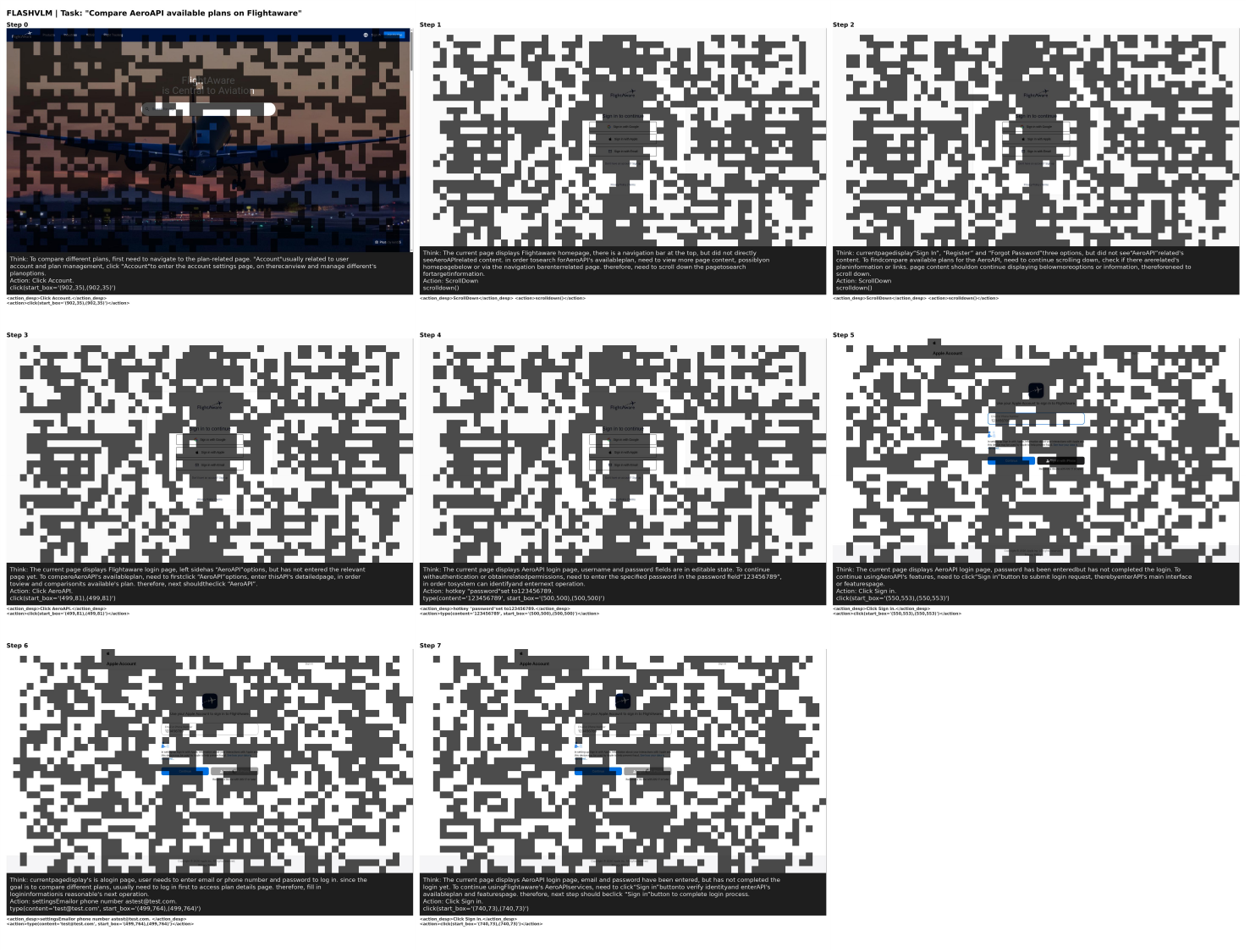}
    \caption{Step-wise visualization of retained visual tokens for \textbf{FlashVLM} over a complete GUI interaction episode at 50 \% keep ratio. Each panel corresponds to one execution step in the same task. The retained-token pattern appears spatially fragmented across steps, which may weaken the local structural and contextual evidence needed for GUI grounding. In this example, the method does not complete the task within 10 steps.}
    \label{fig:flashvlm_fragmentation}
\end{figure*}

\begin{figure*}[t]
    \centering
    \includegraphics[width=\textwidth]{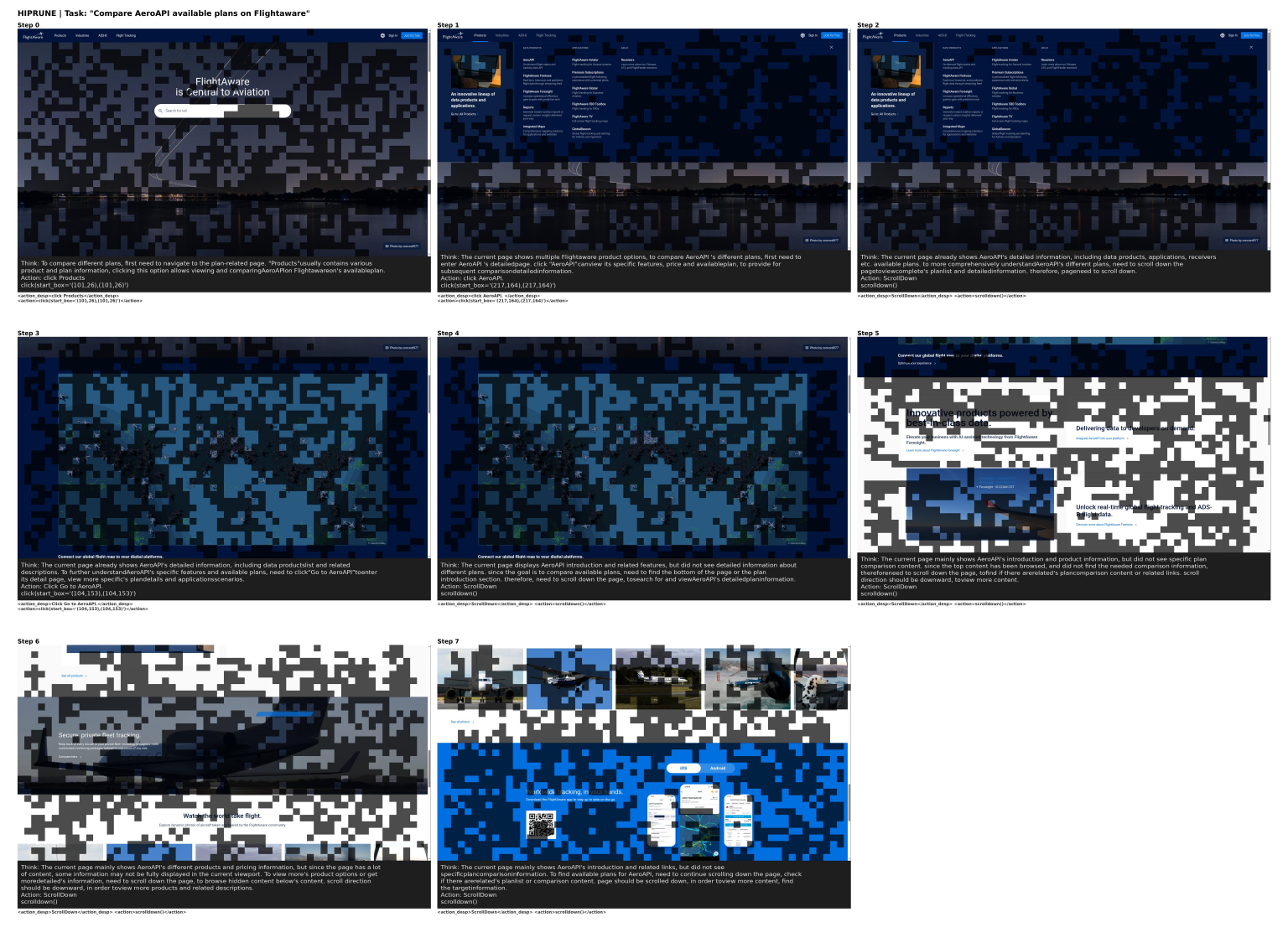}
    \caption{Step-wise visualization of retained visual tokens for \textbf{HiPrune} over a complete GUI interaction episode under a fixed token budget. Each panel corresponds to one execution step in the same task. In this example, the trajectory includes multiple scrolling steps and does not reach task completion.}
    \label{fig:hiprune_fragmentation}
\end{figure*}

\begin{figure*}[t]
    \centering
    \includegraphics[width=\textwidth]{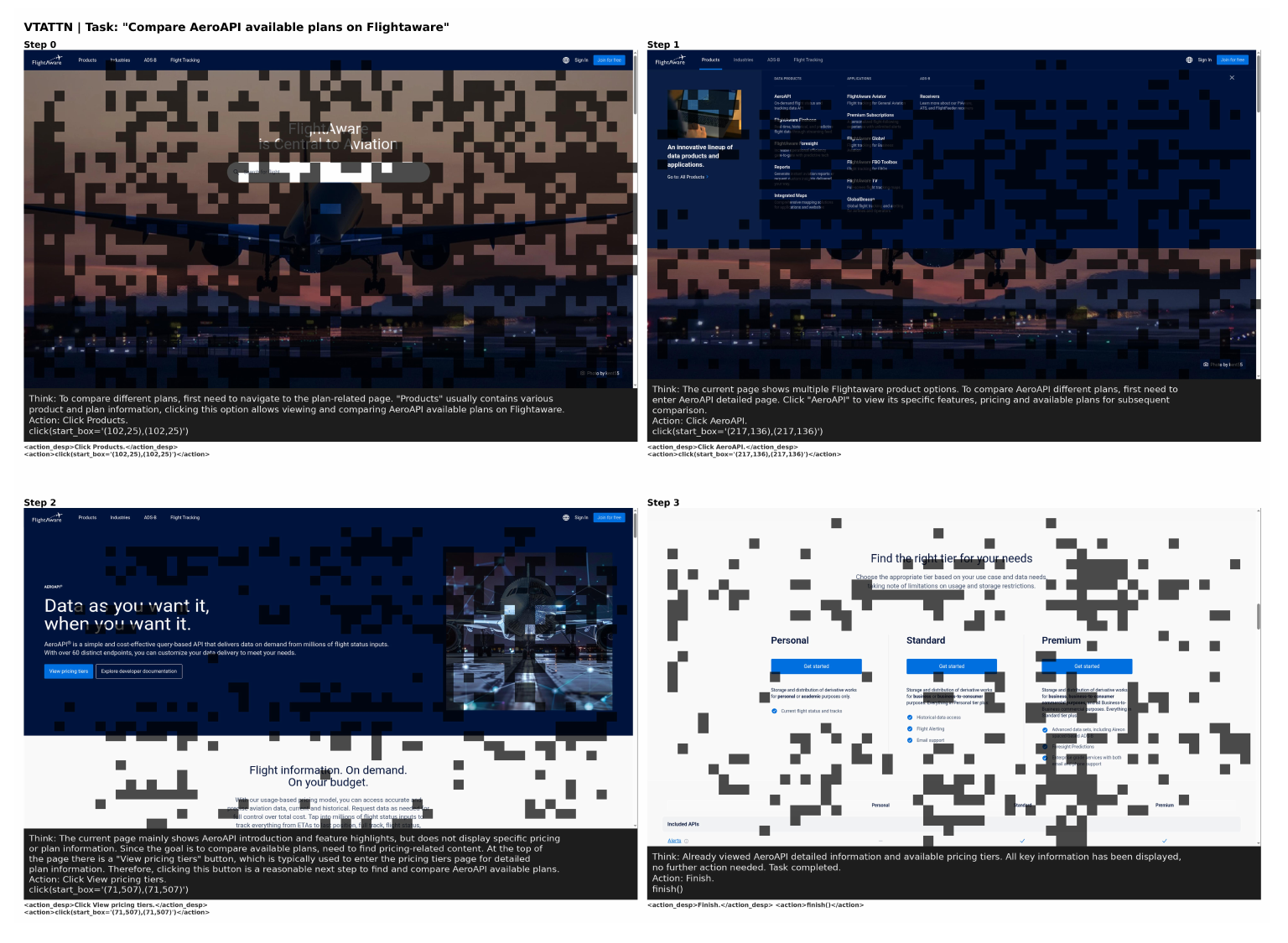}
    \caption{Step-wise visualization of retained visual tokens for \textbf{Attention} over a complete GUI interaction episode at 50 \% keep ratio. Each panel corresponds to one execution step in the same task. This method completes the task in this example, but its retained-token pattern remains more fragmented than that of STP-GUI, while appearing more continuous than some other baselines.}
    \label{fig:vtattn_fragmentation}
\end{figure*}

\begin{figure*}[t]
    \centering
    \includegraphics[width=\textwidth]{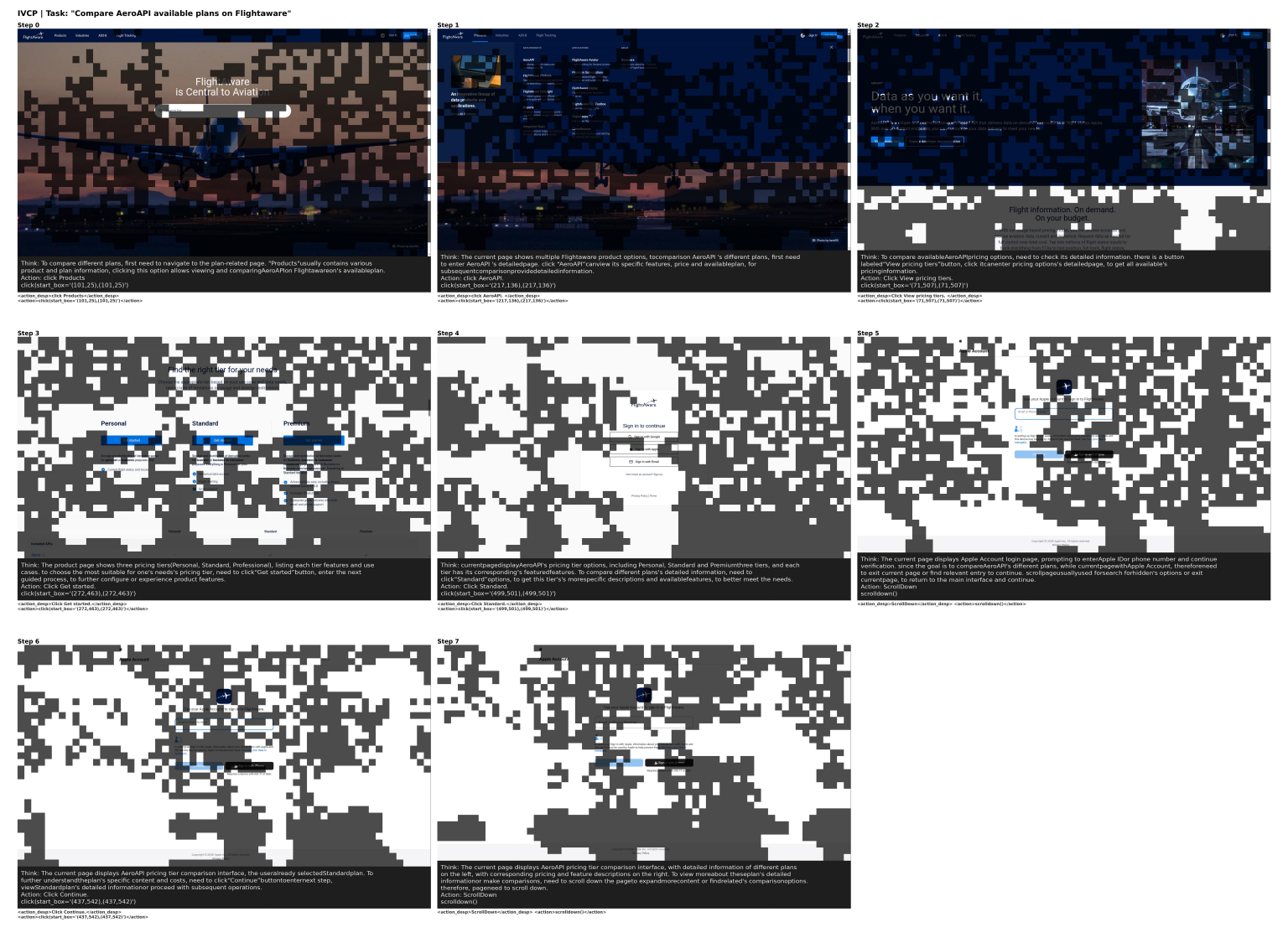}
    \caption{Step-wise visualization of retained visual tokens for \textbf{IVC} over a complete GUI interaction episode at 50 \% keep ratio. Each panel corresponds to one execution step in the same task. In this example, the method executes the first two steps correctly, but later steps exhibit less continuous retained evidence over the webpage, and the task is not completed.}
    \label{fig:ivcp_fragmentation}
\end{figure*}

\begin{figure*}[t]
    \centering
    \includegraphics[width=\textwidth]{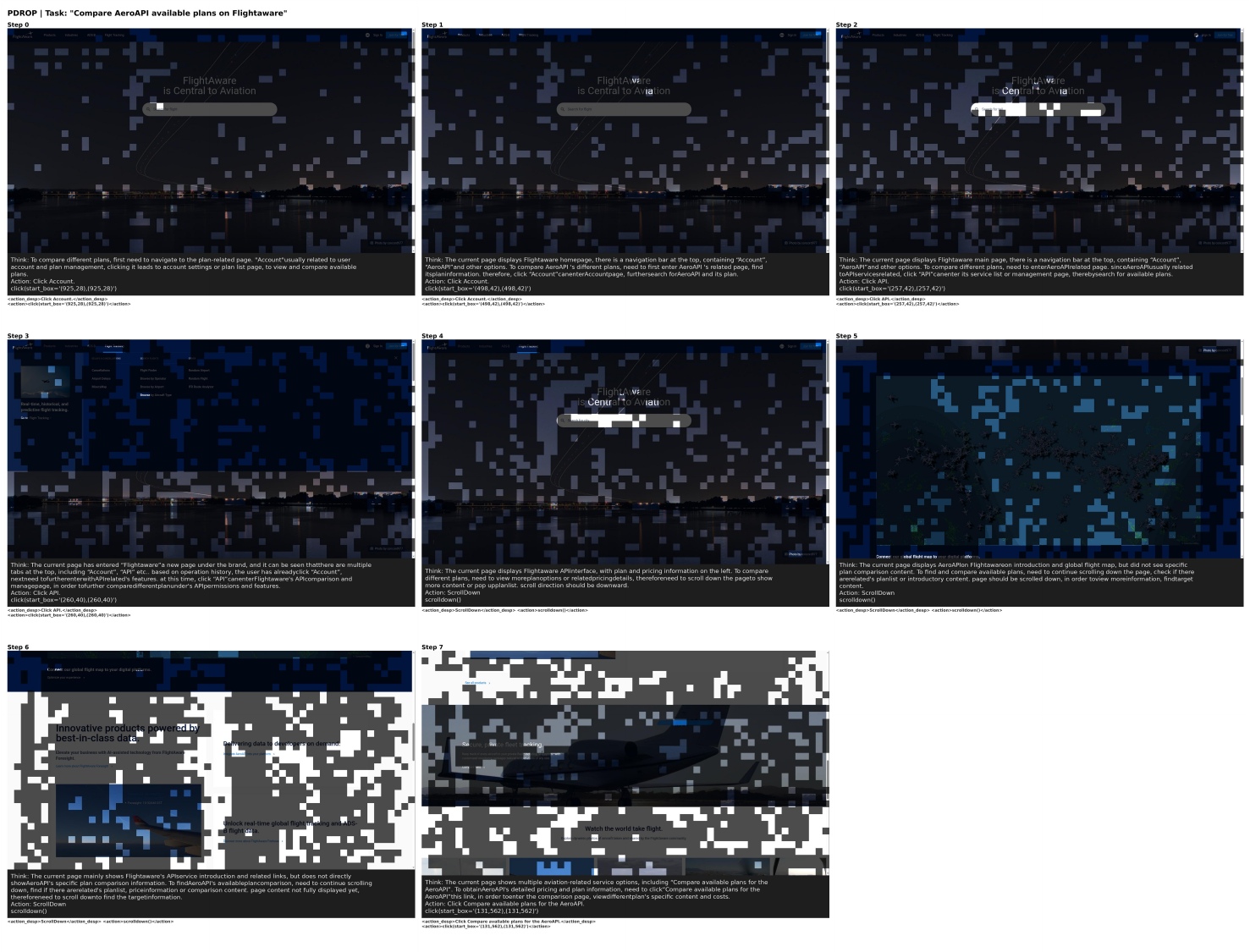}
    \caption{Step-wise visualization of retained visual tokens for \textbf{PDrop} over a complete GUI interaction episode at 41 \% keep ratio. Each panel corresponds to one execution step in the same task. The retained-token pattern appears spatially fragmented across steps, which may weaken the local structural and contextual evidence needed for GUI grounding. In this example, the trajectory returns to the initial page and does not complete the task.}
    \label{fig:pdrop_fragmentation}
\end{figure*}

To further investigate how different pruning strategies affect the spatial structure of retained visual evidence, we provide additional visualizations for 5 representative methods: FlashVLM, PDrop, HiPrune, Attention, and IVC. Figures~\ref{fig:flashvlm_fragmentation}, \ref{fig:pdrop_fragmentation}, \ref{fig:hiprune_fragmentation}, \ref{fig:vtattn_fragmentation}, and \ref{fig:ivcp_fragmentation} show the retained-token patterns of these methods across all execution steps of the same GUI interaction episode(In fact, this example task can be completed in just 4 steps.). Instead of presenting only a single-step snapshot, each large figure visualizes the full step-by-step pruning behavior of one method throughout the entire interaction process.

These full-process visualizations are useful for examining whether a pruning strategy preserves spatially coherent local evidence or gradually degenerates into fragmented token retention over time. This distinction is particularly important for GUI grounding, where accurate action prediction often depends not only on the target region itself, but also on nearby contextual cues such as surrounding text, icons, and local layout structure. As a result, methods that retain visually scattered fragments may fail to preserve the structured local evidence needed for reliable grounding, even when using a similar token budget.

Overall, these visualizations provide complementary qualitative evidence to the quantitative results in the main paper. By comparing the retained-token patterns across multiple execution steps, we can more clearly observe the different fragmentation behaviors induced by different pruning strategies and better understand how preserving local spatial coherence may benefit GUI agents.

\end{document}